\documentclass[journal]{IEEEtran}

\usepackage[nocompress]{cite}
\usepackage[utf8]{inputenc} 
\usepackage[T1]{fontenc}    
\usepackage{hyperref}       
\usepackage{url}            
\usepackage{booktabs}       
\usepackage{amsfonts}       
\usepackage{nicefrac}       
\usepackage{microtype}      
\usepackage{multirow}
\usepackage{graphicx}
\usepackage[caption=false]{subfig}
\usepackage{xcolor}
\usepackage{comment}
\usepackage{amsmath,amssymb}
\usepackage{overpic}
\usepackage{ragged2e}

\graphicspath{{./Imgs/}}
\DeclareGraphicsExtensions{.pdf,.jpg,.png}

\newcommand{\figref}[1]{Fig.~\ref{#1}}
\newcommand{\tabref}[1]{Tab.~\ref{#1}}
\newcommand{\secref}[1]{Section~\ref{#1}}

\newcommand{\equref}[1]{Equ. (\ref{#1})}

\def\ie{\emph{i.e.}}

\newcommand{\add}[1]{#1}

\begin{document}
%
\title{Rethinking 3D LiDAR Point Cloud Segmentation}

\author{Shijie Li,   
        Yun Liu,     
        Juergen Gall 
\thanks{S. Li and J. Gall are with Bonn University, Germany. Y. Liu is with ETH Zurich, Switzerland. S. Li (lishijie@iai.uni-bonn.de) is the corresponding author.}
\thanks{This work was funded by the Deutsche Forschungsgemeinschaft (DFG, German Research Foundation) under Germany's Excellence Strategy - EXC 2070 - 390732324 and GA1927/5-2 (FOR 2535 Anticipating Human Behavior).}%
\thanks{This paper is submitted to IEEE TNNLS Special Issue on Effective Feature Fusion in Deep Neural Networks.}
\thanks{Manuscript received Dec 02, 2020; revised Jun 17, 2021.}}

\markboth{Journal of \LaTeX\ Class Files,~Vol.~14, No.~8, August~2015}%
{Shell \MakeLowercase{\textit{et al.}}: Bare Demo of IEEEtran.cls for IEEE Journals}

\maketitle

\begin{abstract}
Many point-based semantic segmentation methods have been designed for indoor scenarios, but they struggle if they are applied to point clouds that are captured by a LiDAR sensor in an outdoor environment. In order to make these methods more efficient and robust such that they can handle LiDAR data, we introduce the general concept of reformulating 3D point-based operations such that they can operate in the projection space. While we show by means of three point-based methods that the reformulated versions are between 300 and 400 times faster and achieve a higher accuracy, we furthermore demonstrate that the concept of reformulating 3D point-based operations allows to design new architectures that unify the benefits of point-based and image-based methods. As an example, we introduce a network that integrates reformulated 3D point-based operations into a 2D encoder-decoder architecture that fuses the information from different 2D scales. We evaluate the approach on four challenging datasets for semantic LiDAR point cloud segmentation and show that leveraging reformulated 3D point-based operations with 2D image-based operations achieves very good results for all four datasets.
\end{abstract}

\begin{IEEEkeywords}
Semantic segmentation, LiDAR sensor, autonomous driving, point cloud
\end{IEEEkeywords}

\IEEEpeerreviewmaketitle

\section{Introduction}
\IEEEPARstart{E}{nvironment} understanding is essential for autonomous driving. \add{For this goal, the cars are equipped with many sensors and each sensor can be used for different tasks. For example, RGB cameras capture appearance information in order to recognize different objects \cite{9340008}, but they do not provide any depth information. Radar sensors are suitable to measure distance and relative motion and help understanding dynamic scenes \cite{8911477}, but they do not detect small objects. Light detection and ranging (LiDAR) sensors are usually used to capture the environment due to its accurate measurement and the semantic segmentation of the point clouds captured by LiDAR sensors is an essential step for autonomous vehicles. The different sensors complement each other and multi-modal data is commonly used for different tasks like object detection \cite{8937825}, object tracking \cite{9363012}, or semantic segmentation \cite{9000872}.}

\begin{figure*}[!tbp]
\centering
\includegraphics[width=\linewidth]{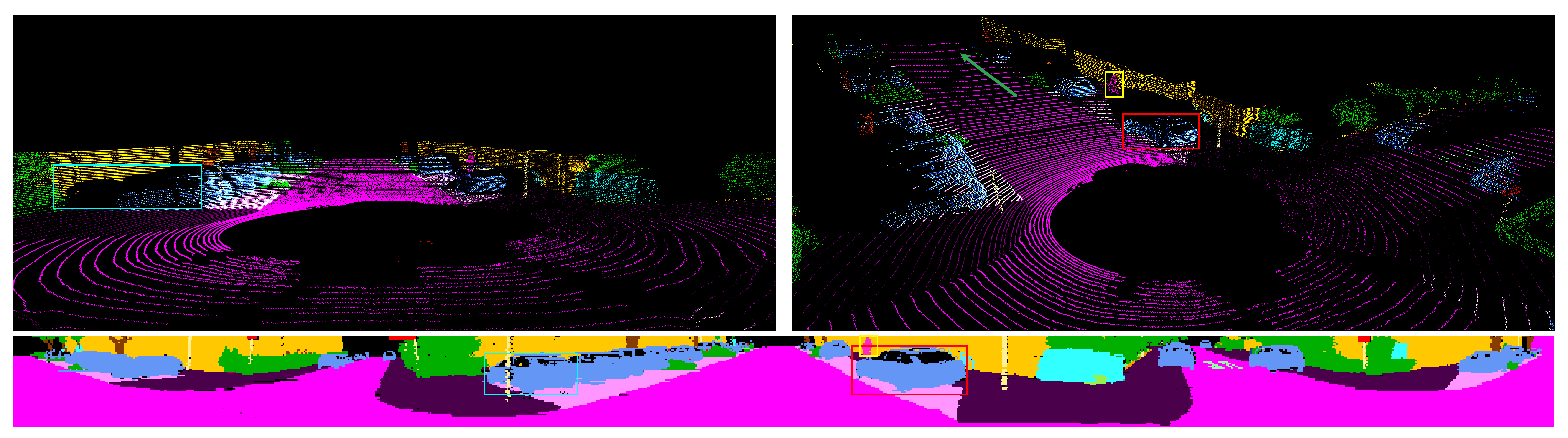}
\vspace{-5mm}
\caption{
Due to the nature of LiDAR sensors, the captured 3D point cloud can be projected onto a plane. This means that neighboring points in 3D are also close in the projected plane, but neighbors in the projected plane can be far distant in 3D. Furthermore, the points become more sparse as the distance to the sensor increases (green arrow), while the points are dense in the projection. We highlight some objects by bounding boxes as an example. Best seen using the zoom function of a PDF viewer.
}
\label{fig:incons}
\vspace{-2mm}
\end{figure*}

In recent years, several deep learning approaches have been proposed that operate on point clouds~\cite{qi2017pointnet,qi2017pointnet++,tatarchenko2018tangent,wu2019pointconv}. 
These point-based methods perform very well for small-scale indoor scenarios where dense point clouds are generated by fusing data captured by RGB-D sensors. 
It was, however, shown in~\cite{behley2019semantickitti} that these methods do not perform well in terms of efficiency and accuracy for point clouds captured by a rotational LiDAR sensor in outdoor scenarios. 
This is due to two reasons. First, the computational cost of point-based methods increases with the total number of points in a point cloud and LiDAR point clouds for outdoor scenes are very large. Second, the density of LiDAR point clouds drops rapidly with the distance to the LiDAR sensor as shown in the top row of \figref{fig:incons}.

\begin{figure}[!t]
\centering
\includegraphics[width=\linewidth,height=1cm]{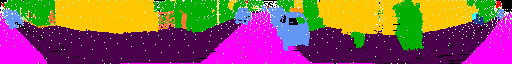}
\\
\vspace{1mm}
\includegraphics[width=\linewidth,height=1cm]{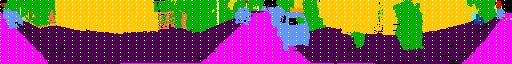}
\caption{Comparison of the sampling methods in PointNet++ (top) and reformulated PointNet++ (bottom). 1024 points are sampled from the LiDAR point cloud. For better visualization, we show the 2D projected image with white points denoting the sampled points. PointNet++ only samples a few points for close objects like cars and most sampled points lie on the distant region where the real distribution of points is sparse.
}
\label{fig:sampling}
\end{figure}

In this work, we address these issues and demonstrate that point-based methods can be reformulated such that they are suitable for LiDAR data. The core idea is that we make use of a projection of the LiDAR point cloud as shown in the bottom row of \figref{fig:incons}. 
In contrast to methods~\cite{wu2018squeezeseg,wu2019squeezesegv2,milioto2019rangenet++} that apply 2D convolutions, we preserve the architectures and the operations of point-based methods. 
Point-based methods comprise several steps that are repeated within the network architecture. 
These steps include the sampling of 3D points of the point cloud, grouping neighboring points for each sampled point, and computing a feature based on the grouped points. 
In this work, we show how these operations can be performed in the projection space and how these operations can be efficiently implemented. 
Although the operations are the same, the projection leads to significant differences. 
For instance, the sampled points are differently distributed as shown in \figref{fig:sampling}. 
The sampled points of the projected-point version are actually better distributed than the sampled points of the point-based methods that oversample the sparse distant points in a LiDAR point cloud. 

We demonstrate the general concept of reformulating point-based methods by means of the three point-based methods PointNet++~\cite{qi2017pointnet++}, SpiderCNN~\cite{xu2018spidercnn}, and PointConv~\cite{wu2019pointconv} and show that the reformulated versions are between 300 and 400 times faster and increase the mIoU by 58\% - 68\%. While the reformulated versions preserve the operations of the original point-based methods, we also demonstrate that the concept of reformulating point-based methods can also be used to develop new architectures that leverage reformulated 3D point-based operations with 2D image-based operations. As an example of such a network, we propose a network for 3D LiDAR point cloud segmentation, which we term Unprojection Network ({UnPNet}). It integrates the reformulated feature propagation of PointConv for  up- and down-sampling into a 2D encoder-decoder architecture. In this way, we exploit 3D operations that are reformulated to operate in the projection space as well as 2D operations that fuse the information from different 2D scales. Furthermore, we employ edge supervision which would be impossible for point-based methods.

We evaluate UnPNet and the reformulated versions of PointNet++~\cite{qi2017pointnet++}, SpiderCNN~\cite{xu2018spidercnn}, and PointConv~\cite{wu2019pointconv} on the SemanticKITTI dataset~\cite{behley2019semantickitti}, which is a large-scale dataset for semantic segmentation of LiDAR point clouds.
Apart from SemanticKITTI, we also evaluate the proposed UnPNet on three other datasets for a comprehensive comparison.
The experiments show that UnPNet performs very well on all four datasets and that it outperforms the reformulated point-based methods since it combines 3D point operations with 2D fusion techniques.  

In summary, we show in this works that
\begin{itemize}
    \item point-based methods can be reformulated such that they operate in the projection space for processing LiDAR point clouds;
    \item the reformulated point-based methods are more efficient and achieve a higher accuracy than the original point-based methods;
    \item the combination of reformulated 3D point-based operations with 2D image-based operations unifies the benefits of point-based and image-based methods.
\end{itemize}
Code will be released at \url{https://github.com/sj-li/UnpNet}.

\begin{figure*}[!t]
\centering
\includegraphics[width=\linewidth]{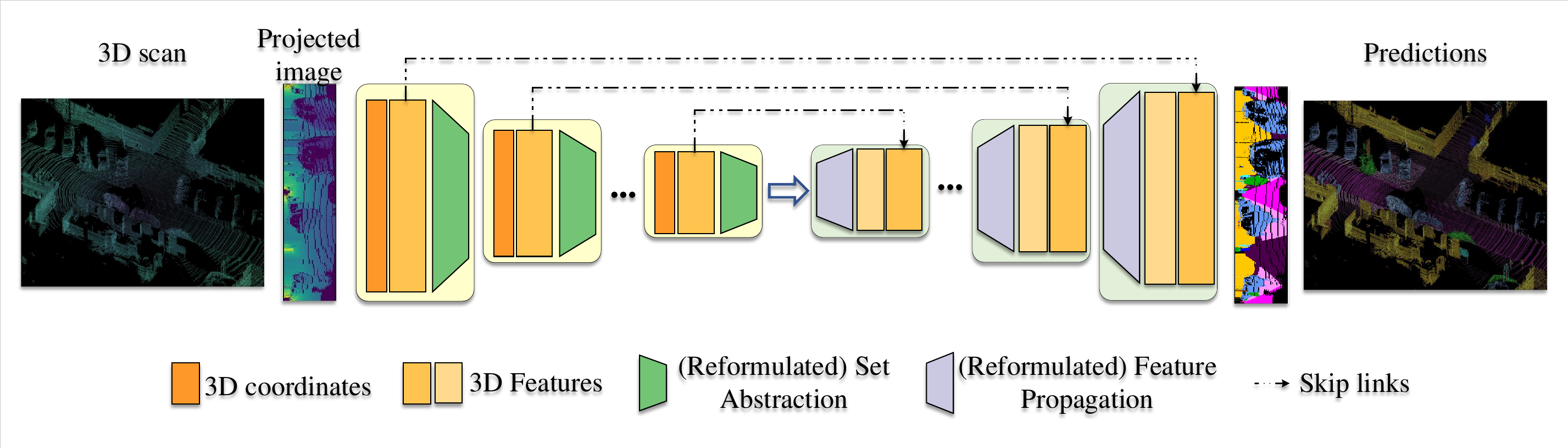}
\vspace{-6mm}
\caption{Overall architecture of PointNet++ or its reformulation (reformulated PointNet++).}
\label{fig:arch}
\vspace{-3mm}
\end{figure*}

\section{Related Work} \label{sec:related}
In this section, we briefly review recent methods for LiDAR point cloud segmentation.
Although CNN-based methods have been very successful for 2D image segmentation, they cannot be directly applied to point cloud segmentation since point clouds do not have the grid structure of images.
To handle this problem, permutation-invariant operations are adopted in PointNet~\cite{qi2017pointnet} to aggregate information, but the method does not capture local structures. This is addressed in PointNet++~\cite{qi2017pointnet} by gathering local information gradually.
The gathering operations in PointNet~\cite{qi2017pointnet} and PointNet++~\cite{qi2017pointnet} are pooling operations which ignore the relative position of 3D points in the local area.
SpiderCNN~\cite{xu2018spidercnn} thus models this information by a polynomial.
Since the distribution of 3D points is usually unbalanced in the 3D space,
PointConv~\cite{wu2019pointconv} explicitly fuses density information into the architecture to improve the representation ability of the model.
While PointCNN~\cite{li2018pointcnn} proposes a generalization of typical CNNs for feature learning on point clouds, 3DMV~\cite{dai20183dmv} combines 2D and 3D features together for better predictions.
Apart from directly processing 3D information, TangentConv~\cite{tatarchenko2018tangent} projects local points to a tangent plane and applies 2D convolutions on it.
The above methods are mainly designed for small-scale scenes with a limited number of points, especially for indoor scenarios.
Different from them, SPGraph~\cite{landrieu2018large} is more suitable for large-scale scenes by defining superpoints to extract compact representations.
These methods, however, do not take the characteristic of the distribution of LiDAR point clouds into consideration and are thus suboptimal in both accuracy and efficiency.

Recently, there are some methods that convert the 3D point cloud to a 2D image according to the scan pattern of LiDAR sensors such that image-based methods can be applied on it. Previous image-based methods are aiming at RGB images.
FCN \cite{long2015fully} treats this task as a dense prediction task and predicts the class probability of each pixel by a fully convolutional network.
Other approaches like \cite{ronneberger2015u} use an encoder-decoder architecture. Although these methods achieve a good performance, they are limited by small receptive fields. To address this problem, DeepLab \cite{chen2014semantic} and its following works \cite{chen2017deeplab,chen2017rethinking,chen2018encoder} introduced dilated convolutions to obtain a larger receptive field and capture image context at
multiple scales.
PSPNet \cite{zhao2017pyramid} proposed a pyramid pooling module to extract contextual information. Due to the importance of semantic segmentation for autonomous driving, some works focus on this area like DenseASPP \cite{yang2018denseaspp}. 
As an comparison, \cite{qiu2020miniseg,liu2018deep} aim at other directions.
These image-based methods, however, are not designed for processing and fusing multi-modal LiDAR data.

To better fit to the application of autonomous driving, some projection-based methods have been proposed. Compared to image-based methods, they are more suitable to process projections of LiDAR data and hence achieve a good accuracy while maintaining a high efficiency.
FuseSeg~\cite{krispel2020fuseseg} combines color and spatial information to segment LiDAR point clouds.
DeepTemporalSeg~\cite{dewan2019deeptemporalseg} proposes a temporally consistent method for LiDAR point cloud segmentation.
SqueezeSeg~\cite{wu2018squeezeseg,wu2019squeezesegv2} uses SqueezeNet~\cite{iandola2016squeezenet} as backbone and a conditional random field (CRF) for post-processing.
PointSeg~\cite{wang2018pointseg} uses a similar architecture as SqueezeNet~\cite{iandola2016squeezenet}, but uses dilated convolutions to increases the receptive field.
Based on SqueezeSeg, RangeNet++~\cite{milioto2019rangenet++} replaces the backbone with Darknet~\cite{redmon2018yolov3} and uses $k$-Nearest-Neighbor ($k$-NN) search for post-processing.
\cite{8580596, 8351244} are projection-based methods designed for detecting drivable regions. While they have been implemented on FPGAs and are very efficient, they do not recognize other semantic objects.
\cite{9134888,akadas20203d,10.1007/978-3-030-64559-5_16,li2020multi} are projection-based methods, which also achieve a good accuracy and are very efficient. 
Some related applications also utilize projection-based methods, like moving object segmentation \cite{chen2021moving}.
While projection-based methods are efficient, nearby points in the projected image can be far away in the 3D space as shown in \figref{fig:incons}. In this work, we therefore explore the inherent relation between projection-based and point-based methods that preserve the 3D structure.


\section{Reformulation of Point-based methods} \label{sec:reform}
In order to show how a network operating on LiDAR points can be reformulated to operate in the projection space, we use PointNet++~\cite{qi2017pointnet++} as an example. In Section~\ref{sec:other}, we discuss the reformulated examples of two other point-based networks, namely SpiderCNN~\cite{xu2018spidercnn} and PointConv~\cite{wu2019pointconv}. Before we discuss our approach in \secref{sec:PPNet}, we briefly discuss the main operations of PointNet++.

\subsection{Review of PointNet++}\label{sec:rev}
We choose PointNet++~\cite{qi2017pointnet++} as an example since it is very popular and has been used as the baseline in many works.
The pipeline of PointNet++ is shown in \figref{fig:arch}.
PointNet++ consists of so-called set abstraction modules and feature propagation modules as shown in \figref{fig:arch}. 
The set abstraction module comprises a sampling layer, a grouping layer, and a PointNet layer. The sampling layer chooses a subset from the input point set, which defines the centroids of local regions. \figref{fig:sampling} shows the sampled points. The grouping layer groups the neighboring points of each centroid, which forms a local region. The PointNet layer computes a feature vector based on the neighboring points using a multilayer perceptron (MLP) and max pooling. 
While the set abstraction modules subsample the original point set, the feature propagation modules recover the original point set by distance based interpolation: 
\begin{equation} \label{equa:interpolation}
    f^{(j)}(x)= \frac{\sum_{i=1}^{k} w_{i}(x) f_{i}^{(j)}}{\sum_{i=1}^{k} w_{i}(x)},
\end{equation}
\begin{equation}
    w_{i}(x)=\frac{1}{d\left(x, x_{i}\right)^{p}}, j=1, \ldots, C, 
\end{equation}
where $f$ is a point-wise feature and $d(x_i, x_j)$ is the distance between point $x_i$ and $x_j$.

\begin{figure}[t]
    \centering
    \subfloat[Farthest point sampling]{
        \includegraphics[width=0.23\textwidth]{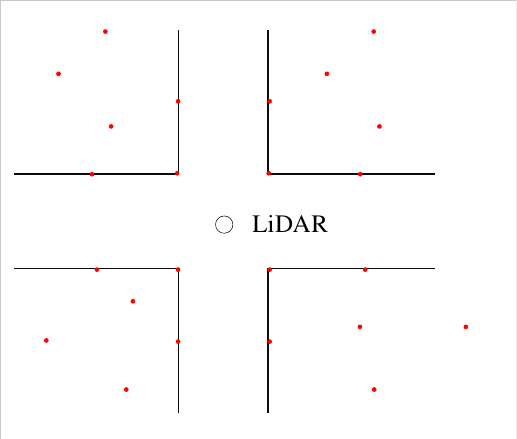}
    }
    \subfloat[Ray sampling]{
        \includegraphics[width=0.2\textwidth]{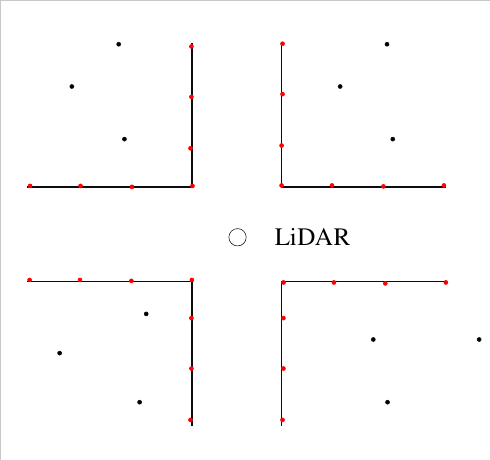}
    }
    \caption{Toy example. The points on the black lines are correct measurements while the other points are outliers. Farthest point sampling selects the outliers such that the sampled points are uniformly distributed in the 3D space. The proposed ray sampling only selects the points (red) on the black lines. The outliers (blue) are not selected.  }
    \label{fig:sampling_toy}
\end{figure}

\subsection{Reformulated PointNet++}\label{sec:PPNet}

To reformulate PointNet++ so that it operates in the projection space, we will not change the architecture, but we need to reformulate the set abstraction and the feature propagation module as shown in \figref{fig:arch}. 
As input, we use the projected LiDAR point cloud as it has been proposed in \cite{wu2018squeezeseg}. 
The projection map is obtained from the LiDAR point cloud by
\begin{equation}\label{eq:proju}
    u = \frac{1}{2}[1 - \arctan(y, x)\pi^{-1}]w,
\end{equation}
\begin{equation}\label{eq:projv}
    v = [1 - (\arcsin(zr^{-1}) + o_{up})o^{-1}]h,
\end{equation}
where $(u, v)$ are the coordinates in the projection map with size $(h, w)$ and $\mathbf{a}=(x, y, z)$ are the 3D coordinates of the points. While $r$ is the depth of the point, $o = o_{up} + o_{down}$ is the vertical field-of-view of the LiDAR sensor.

We will first describe the sampling layer (\secref{sec:Sample}), the grouping layer (\secref{sec:Group}), and the PointNet layer (\secref{sec:Point}) of the reformulated set abstraction and then discuss the reformulated feature propagation (\secref{sec:Prop}).

\begin{figure}[!t]
    \centering
    \includegraphics[width=\linewidth]{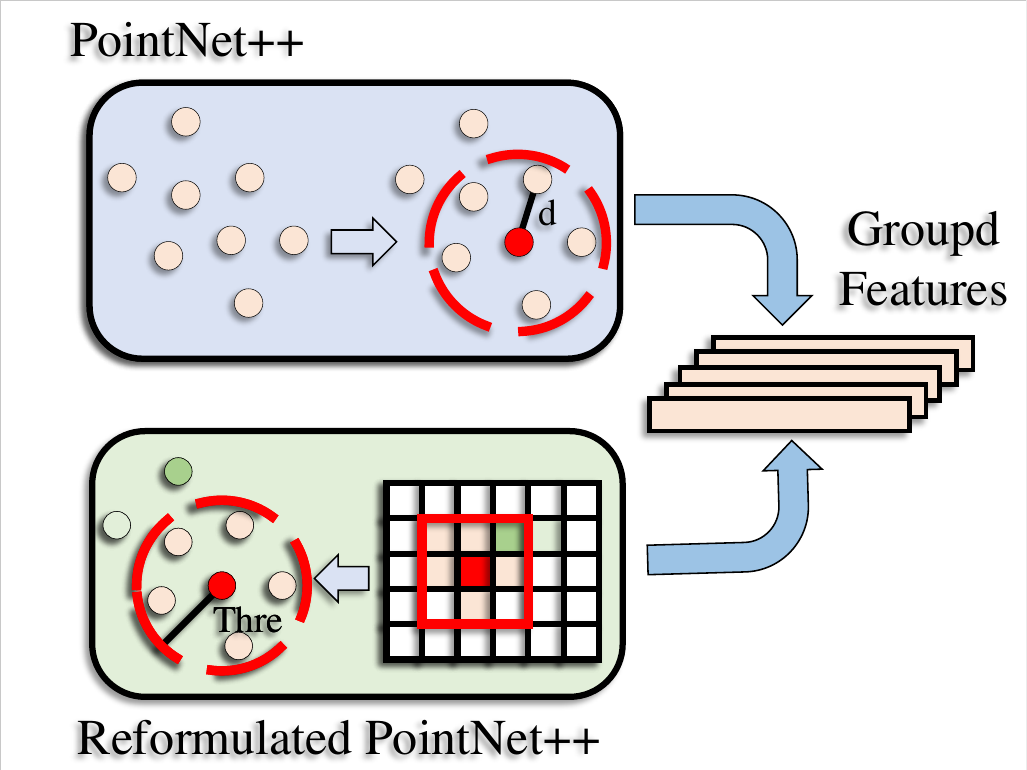}
    \caption{Comparison of the reformulated grouping layer (bottom) with the original one in PointNet++~\cite{qi2017pointnet++} (top). PointNet++ uses a ball query to obtain all neighboring points (orange) within a certain radius for each sampled point (red). In contrast, our method first searches the $k \times k$ neighboring rays, and then we discard the points that are outside the radius (dark green point).}
    \label{fig:grouping_comp}
    \vspace{-3mm}
\end{figure}

\subsubsection{Reformulated Sampling Layer}\label{sec:Sample}
PointNet++ \cite{qi2017pointnet++} uses farthest point sampling to sample a subset of 3D points. 
It is designed to maximize the distance between sampled points that are thus uniformly scattered in the 3D space.
However, the real distribution of LiDAR points is not uniform and becomes sparse as the distance to the sensor increases as shown in \figref{fig:incons}.
This mismatch harms the performance when applying farthest point sampling to LiDAR points, as shown in \figref{fig:sampling}.
Furthermore, the computational complexity of farthest point sampling is $\mathcal{O}(N\log N)$ where $N$ is the number of 3D points~\cite{kamousi2016analysis}. This makes the approach highly inefficient for large point clouds which are common for LiDAR sensors.
Therefore, farthest point sampling is suboptimal for LiDAR point cloud segmentation in terms of both effectiveness and efficiency.
To address this problem, we propose to uniformly sample the 3D points from the projected point cloud as shown in \figref{fig:sampling}. 
This has the advantage that we sample rays instead of points, which means that the distance between the sampled points is larger if they are farther away from the sensor. 
Hence, the distribution of sampled points accords with the original point cloud.
\add{The sampling is also less sensitive to outliers. Since farthest point sampling aims to sample points that are uniformly in the 3D space, it tends to select all outliers that are distant to correct measurements. In case of ray sampling, the probability to select an outlier is equivalent to the percentage of outliers and thus lower compared to farthest point sampling as it is illustrated in \figref{fig:sampling_toy}.}
Another benefit is that we can use a 2D grid structure for sampling such that the computational complexity becomes $\mathcal{O}(M)$ where $M = H' \times W'$ is the number of sampled points and $M \ll N$.

\subsubsection{Reformulated Grouping Layer}\label{sec:Group}

For grouping neighboring 3D points, PointNet++ uses a ball query to obtain for each sampled point all points that are within a given distance. 
While a naive implementation has the complexity of $\mathcal{O}(MN)$, more efficient implementations reduce it using data structures like k-d trees or octrees \cite{behley2015efficient}. 
This, however, increases the memory requirements. 
In order to make the grouping of PointNet++ \cite{qi2017pointnet++} much more efficient, we search neighboring rays first and then exclude the points that are too far away from the sampled point. 
We obtain the neighboring rays by taking the $k\times k$ neighbors in the projected point cloud as shown in \figref{fig:grouping_comp}. 
The parameter $k$ provides a trade-off between accuracy and runtime as we will show in the experiments. 
For each of the $k^2$ points, we obtain the 3D points and subtract the 3D position of the sampled point to convert the points from global coordinates to local coordinates as in PointNet++. 
We then compute the norm of each point, \ie, the 3D distance to the sampled point, and mask all points that are within a given distance. 
The complexity of this operation is $\mathcal{O}(Mk^2)$ where $k^2 \ll \log N$. 
A comparison between this grouping strategy and the grouping in PointNet++ is displayed in \figref{fig:grouping_comp}.

\begin{figure}[!t]
\centering
\vspace{-3mm}
\includegraphics[width=\linewidth]{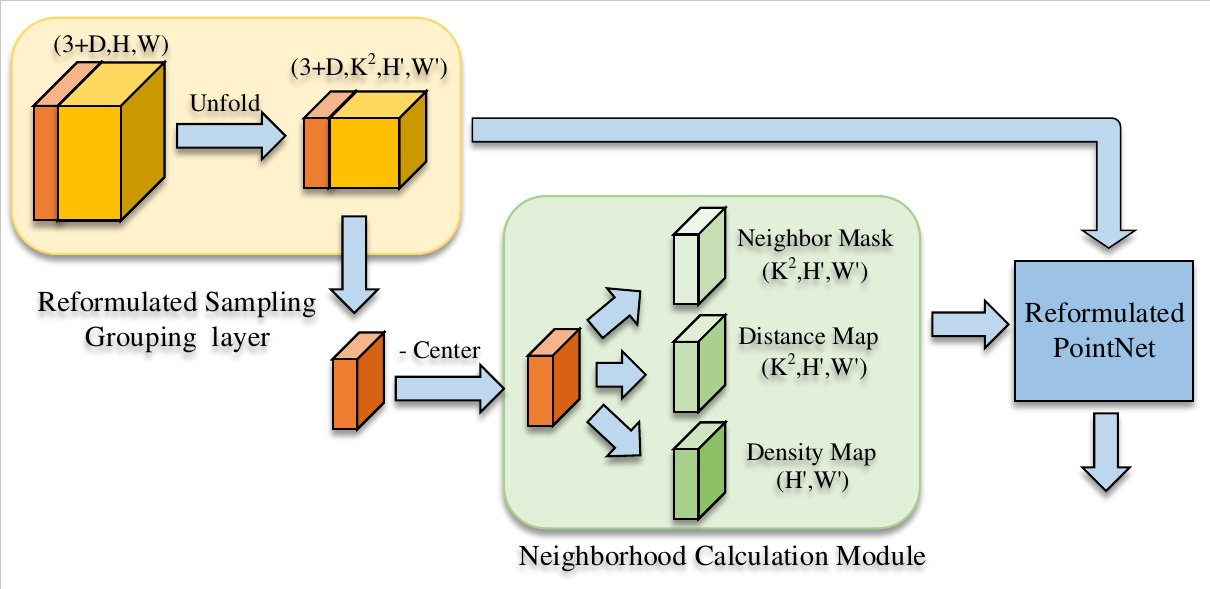}
\caption{Illustration of the reformulated set abstraction.}
\label{fig:ppa}
\end{figure}

\figref{fig:ppa} illustrates how reformulated sampling and grouping are efficiently implemented in a network. 
Given the input $\mathbb{R}^{(C+3)\times H\times W}$, where $C$ is the number of feature channels which are concatenated with the 3D coordinates ($C+3$), the unfold operation uniformly samples $H'\times W'$ points as discussed in \secref{sec:Sample} and copies the corresponding $k\times k$ neighborhood for each sampled point $m \in H'\times W'$. This yields the tensor $F_{in} \in \mathbb{R}^{(C+3)\times k^2\times H'\times W'}$. 
For each sampled point, we then subtract the 3D coordinate of the sampled point from the coordinates of the corresponding $k^2$ neighboring points. 
We finally compute the distance map $\mathbb{R}^{k^2\times H'\times W'}$ and the binary neighborhood mask $\{0,1\}^{k^2\times H'\times W'}$, which is $1$ if a point is within the radius of a sampled point. The neighborhood mask defines the grouping for each sampled point.

At this step, we directly compute the inverse distance map, which will be used for the reformulated feature propagation and will be described in the next section, and the inverse density map as described in \cite{wu2019pointconv}. 
The latter will be needed for converting PointConv \cite{wu2019pointconv} into a reformulated point-based method.

\begin{figure}[!t]
\centering
\begin{overpic}[width=\linewidth,height=2cm]{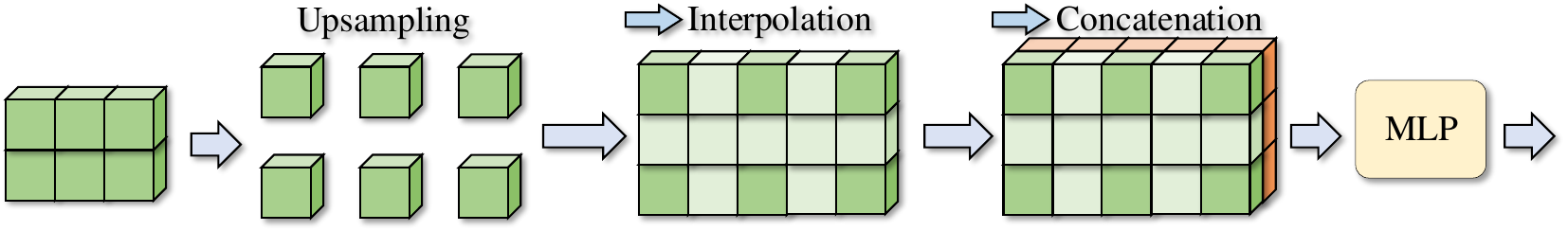}
        \put(36.0,19){$\hat{\mathcal{D}}$}
        \put(59.5,19){$\hat{F}$}
    \end{overpic}
\caption{Illustration of the reformulated feature propagation.}
\label{fig:interpolation}
\vspace{-3mm}
\end{figure}

\begin{figure*}[!t]
\centering
\vspace{-5mm}
\subfloat[PointNet++]{\label{fig:pnet}
\includegraphics[width=.48\linewidth]{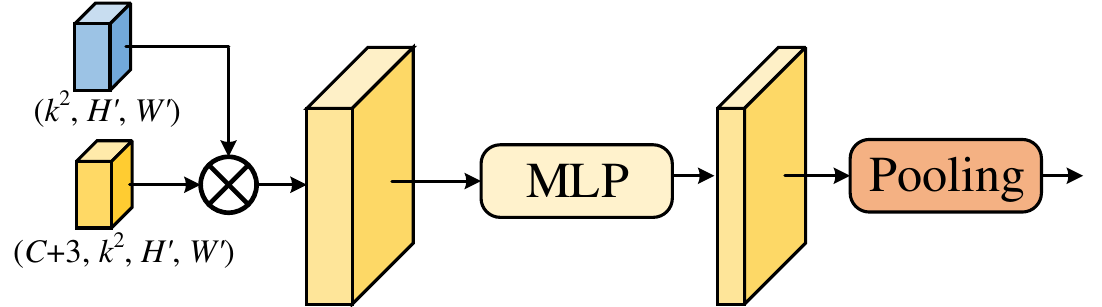}
}
\subfloat[Legend]{
\includegraphics[width=0.48\linewidth]{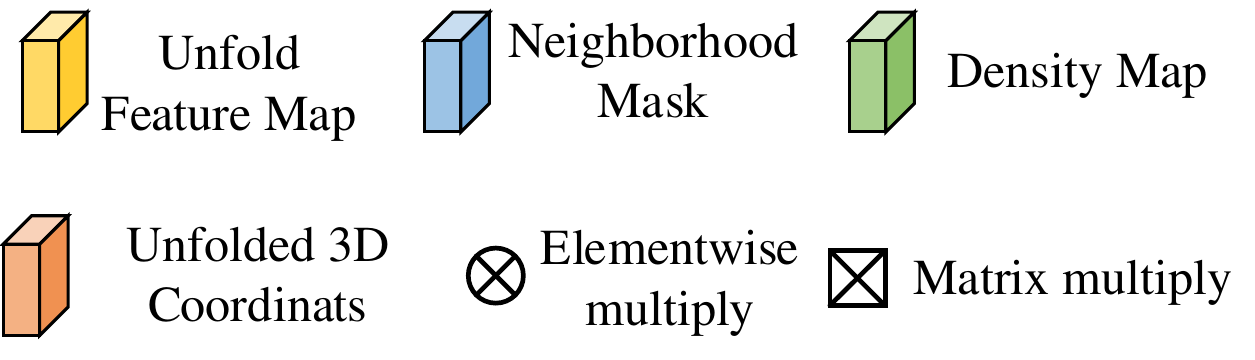}
}
\\ \vspace{-3mm}
\subfloat[SpiderCNN]{\label{fig:scnn}
\includegraphics[width=0.48\linewidth]{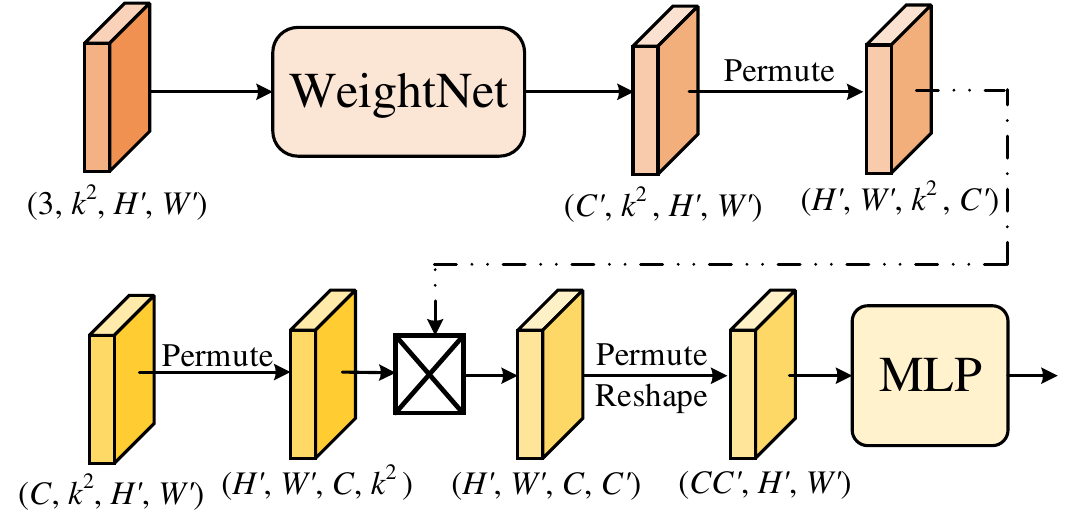}
}
\subfloat[PointConv]{\label{fig:pconv}
\includegraphics[width=0.48\linewidth]{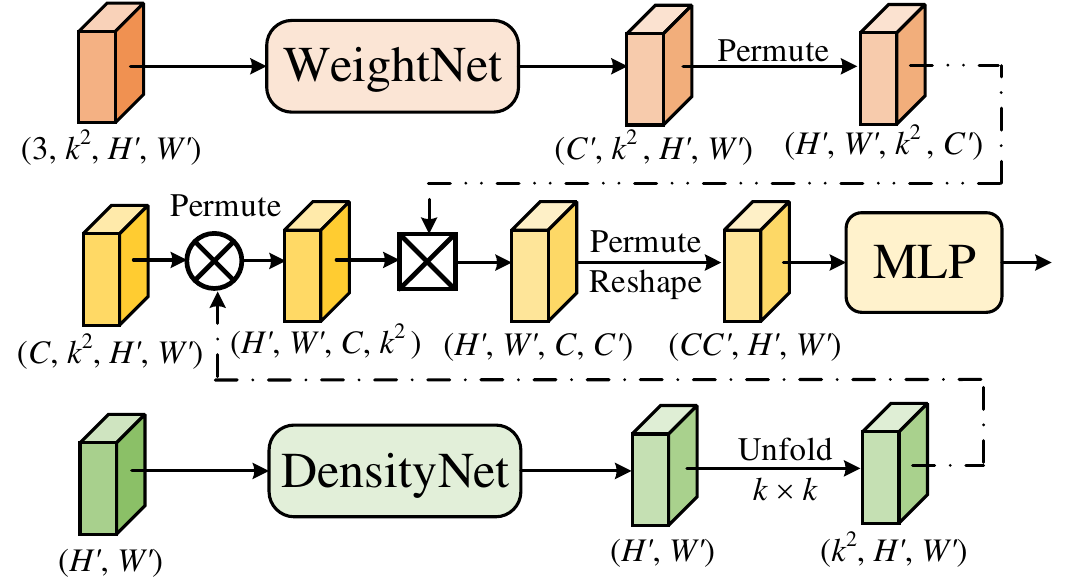}
}
\caption{Reformulation of the set abstraction module for PointNet++~\cite{qi2017pointnet++}, SpiderCNN~\cite{xu2018spidercnn}, and PointConv~\cite{wu2019pointconv}. In case of a dimension mismatch in the element-wise operations, we use the broadcasting mechanism, which is omitted in the illustrations since it is the default operation in modern deep learning frameworks like PyTorch~\cite{paszke2019pytorch}. We also omit ${\rm MLP}_{in}$ from \equref{equa:group_pscnn} and \equref{equ:group_pconv}.}
\label{fig:grouping}
\vspace{-4mm}
\end{figure*}

\subsubsection{Reformulated PointNet Layer}\label{sec:Point}
The PointNet layer in PointNet++ uses max pooling and an MLP. 
Given $F_{in} \in \mathbb{R}^{(C+3)\times k^2\times H'\times W'}$ after the unfold operation and the neighborhood mask, the reformulated PointNet layer can thus be denoted as
\begin{equation} \label{equ:pointnet}
    F_{out} ={\rm Pooling}({\rm MLP}(F_{in}\otimes \mathcal{M})).
\end{equation}
The symbol $\otimes$ denotes element-wise multiplication and $\mathcal{M}$ is the neighborhood mask, which has been duplicated $(C+3)$-times to have the same size as $F_{in}$.
\figref{fig:pnet} illustrates this operation.

\subsubsection{Reformulated Feature Propagation Module}\label{sec:Prop}

As described in \secref{sec:rev} and illustrated in \figref{fig:arch}, the feature propagation modules recover the original point set by the distance based interpolation (\equref{equa:interpolation}). 
Since our sampled points are uniformly distributed on the projected image, this can be very efficiently implemented. 
The sampled points are first set back to its original positions. 
The distance based interpolation (${\rm Interp}$) is then applied as in \equref{equa:interpolation} using the precomputed inverse distance map $\hat{\mathcal{D}}$. 
As illustrated in \figref{fig:arch}, there are skip connections between the blocks. The interpolated features ${\rm Interp}(F_{in}, \hat{\mathcal{D}})$ are thus concatenated with the point features from the corresponding set abstraction module ($\hat{F}$) and fed into an MLP, \ie, 
\begin{equation}
    \label{equa: inter}
    F_{out} = {\rm MLP}({\rm Concat}({\rm Interp}(F_{in}, \hat{\mathcal{D}}), \hat{F})).
\end{equation}
The reformulation of the Feature Propagation module is illustrated in \figref{fig:interpolation}.

\subsection{Reformulated SpiderCNN and PointConv}\label{sec:other}
So far we discussed how PointNet++~\cite{qi2017pointnet++} can be reformulated so that it operates in the projection space, but the approach can be applied to other point-based networks as well. In this section, we therefore briefly describe how two other networks, namely SpiderCNN~\cite{xu2018spidercnn} and PointConv~\cite{wu2019pointconv}, are reformulated. \figref{fig:grouping} illustrates the differences between the reformulation of PointNet++, SpiderCNN, and PointConv.

The reformulated SpiderCNN (RSpiderCNN) can be viewed as a `soft' version of RPointNet++ because it weights each point feature according to its relative position to the corresponding sampled point. 
Instead of using a neighborhood mask, it computes the weight for each point. 
In our case, the operations are performed for the $k \times k$ neighborhood. 
The operations of the set abstraction in RSpiderCNN are thus defined by 
\begin{equation}
\label{equa:group_pscnn}
    F_{out} = {\rm MLP}_{out}({\rm MLP}_{in}(F_{in}) \boxtimes \textit{WeightNet}(\mathcal{P})),
\end{equation}
where the symbol $\boxtimes$ denotes matrix multiplication, and $\mathcal{P}$ denotes the unfolded 3D coordinates after subtracting the 3D coordinates of the corresponding sampled point as shown in \figref{fig:ppa}.
Here, we omit the dimensions which are shown in \figref{fig:scnn}.
For feature propagation, RSpiderCNN follows \equref{equa: inter} and the main difference is that RSpiderCNN applies \equref{equa:group_pscnn} on the interpolated features.

\begin{figure*}[!t]
\centering
\includegraphics[width=\linewidth]{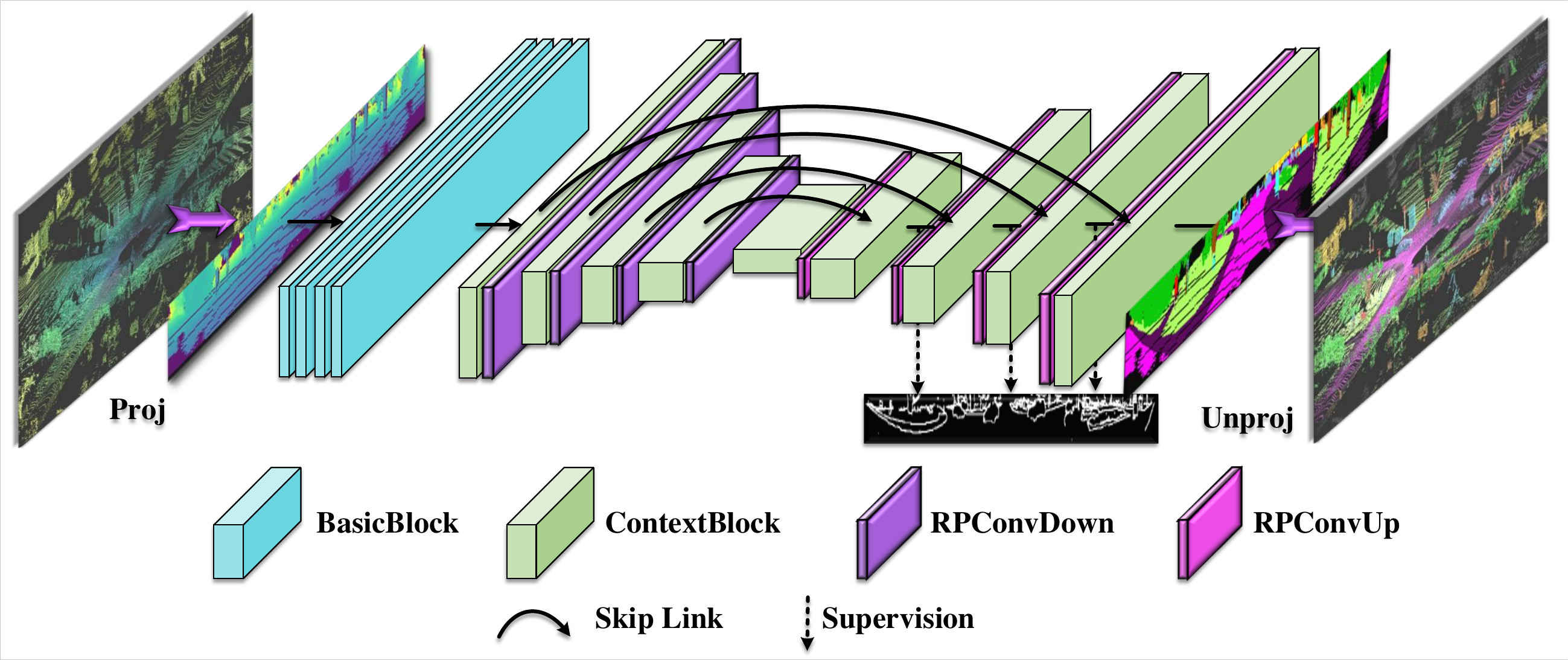}
\vspace{-6mm}
\caption{Overview of UnPNet. The network combines 2D operations like the 
basic block and the context block as well as reformulated 3D operations. In this example, we use the reformulated feature propagation of PointConv for down- and upsampling. The corresponding operations are denoted by RPConvDown and RPConvUp, respectively.}
\label{fig:unpnet}
\vspace{-3mm}
\end{figure*}

\begin{figure}[!t]
\centering
\includegraphics[width=0.8\linewidth]{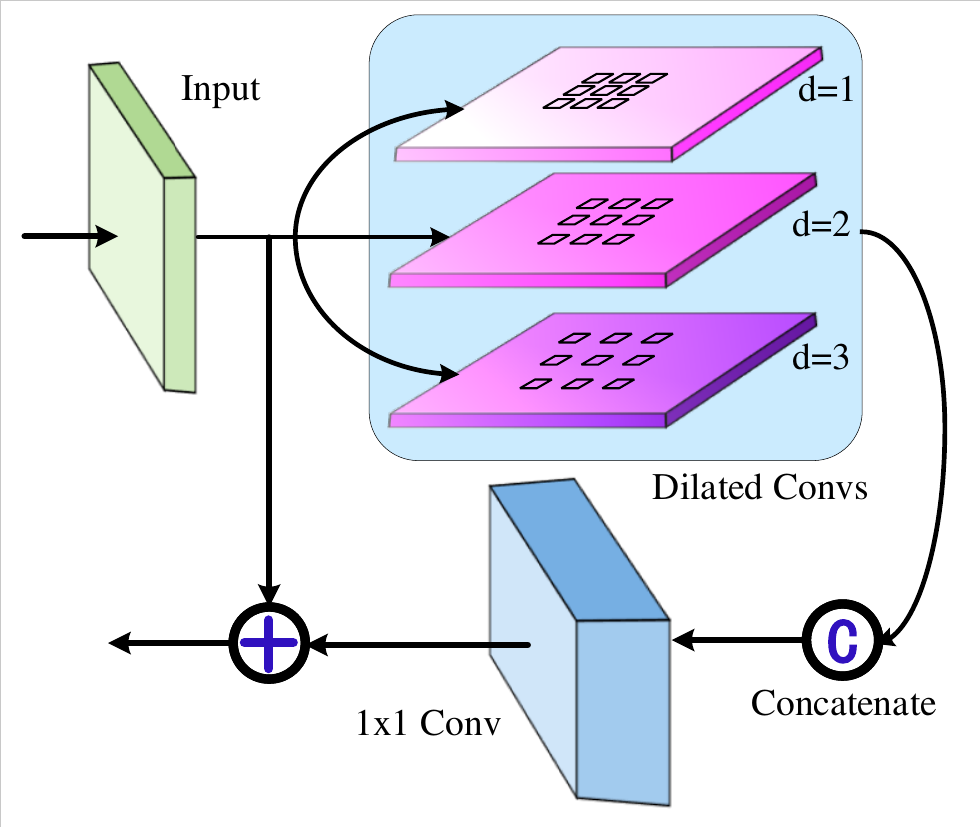}
\caption{Illustration of the context block.}
\label{fig:contextblock}
\vspace{-3mm}
\end{figure}

The reformulated PointConv (RPointConv) uses the point density as additional information. The green branch in \figref{fig:pconv} therefore takes the inverse density map $\mathcal{D}$ from \figref{fig:ppa} as input. 
The operations of the set abstraction in RPointConv are defined by
\begin{equation}
\label{equ:group_pconv}
\begin{aligned}
    F_{out} = & {\rm MLP}_{out}( ({\rm MLP}_{in}(F_{in}) \otimes \\ & \textit{DensityNet}(\mathcal{D})) \boxtimes \textit{WeightNet}(\mathcal{P})).
\end{aligned}
\end{equation}
Similar to RSpiderCNN, RPointConv follows \equref{equa: inter} for feature propagation, but it applies \equref{equ:group_pconv} on the interpolated features.

These examples show that point-based methods can be reformulated to operate in the projection space. In the experiments, we will show that the reformulation makes the point-based methods 300-400 times faster and increases the accuracy.  

\section{Unprojection Network (UnPNet)} \label{sec:unpnet}
So far, we have shown how point-based methods can be reformulated without changing the architecture design and principles. 
The reformulated approaches have the potential to leverage concepts from 3D point-based methods and 2D image-based methods. 
In order to demonstrate this, we propose a network that uses the reformulated feature propagation of PointConv (\equref{equ:group_pconv}) for up- and down-sampling and we integrate it into a 2D CNN with an encoder-decoder architecture. 
In this way, we exploit 3D operations that are reformulated to operate in the projection space as well as 2D operations that fuse the information from different 2D scales.

The network architecture is shown in \secref{sec:PPNet}. 
As in the reformulated architectures, we first project the LiDAR point cloud using \equref{eq:proju} and \equref{eq:projv}. 
Besides the basic block (\ie, 2D convolutions with residual link), the network uses the reformulated feature propagation of PointConv (\equref{equ:group_pconv}) for up- and down-sampling and an additional context block shown in \figref{fig:contextblock} for fusing image context at multiple 2D scales.
In order to make the context block as efficient as possible, we use 2D dilated convolutions. 
More in detail, we use three $3 \times 3$ convolutions with different dilation rates (1, 2, 3) to extract multi-scale features. 
The three feature maps are then concatenated and fused by a $1\times1$ convolution.
In addition, a residual link is employed to facilitate the gradient flow. 
Furthermore, we apply edge supervision to the decoder, which will be described in \secref{sec:supervision}, to ensure better segment boundaries. 
As in \cite{milioto2019rangenet++}, $\textit{k}$-NN can be used for post-processing. 
We call this architecture Unprojection Network (UnPNet), and we will show in the experiments that UnPNet outperforms RPointConv by a large margin. 
While UnPNet is just an example, it demonstrates that the reformulation of point-based methods is a new general concept, allowing to construct new architectures by leveraging 3D point-based networks and 2D CNNs.



\begin{table*}[!t]
    \centering
    \caption{Effect of different parameters.}
    \label{tab:ablation}
    \vspace{-5mm}
    \subfloat[Search size $k$]{
    \resizebox{0.3\linewidth}{!}{
    \begin{tabular}{c|c|c|c}
    \hline
    $k$ & Acc & mIoU & Scans/s \\
    \hline
    3 & 71.4 & 26.8 & 38.2 \\
    \hline
    5 & 74.7 & 30.7 & 30.0 \\
    \hline
    7 & 76.2 & 31.9 & 23.7 \\
    \hline    
    \end{tabular}}
    }
    \subfloat[Input resolution $R$]{
    \resizebox{0.37\linewidth}{!}{
    \begin{tabular}{c|c|c|c}
    \hline
    R & Acc & mIoU & Scans/s \\
    \hline
    64$\times$512 &  74.7 & 30.7 & 30.0 \\
    \hline
    64$\times$1024 & 77.4 & 31.3 & 17.2 \\
    \hline
    64$\times$2048 & 75.2 & 25.7 & 9.4\\
    \hline    
    \end{tabular}}
    }
    \subfloat[Sampling stride $S$]{
    \resizebox{0.3\linewidth}{!}{
    \begin{tabular}{c|c|c|c}
    \hline
        $S$ & Acc & mIoU & Scans/s \\
        \hline
        1 & 74.5 & 30.4 & 22.2 \\
        \hline
        2 & 74.7 & 30.7 & 30.0 \\
        \hline
        4 & 69.6 & 25.0 & 34.5 \\
    \hline    
    \end{tabular}}
    }
\end{table*}

\begin{table*}[!tb]
    \centering
    \setlength{\tabcolsep}{3.2mm}
    \caption{Comparison with the original baseline methods.}
    \label{tab:comparision_origin}
    \vspace{-2mm}
    \resizebox{\linewidth}{!}{%
    \begin{tabular}{c|cc|cc|cc}
    \hline
    & PointNet++~\cite{qi2017pointnet++} & RPointNet++ & SCNN~\cite{xu2018spidercnn} & RSCNN & PointConv~\cite{wu2019pointconv} & RPointConv \\
    \hline
    Acc & 64.6 & 74.7 & 70.3 & 81.5 & 72.5 & 82.5\\
    \hline
    mIoU & 19.4 & 30.7 & 21.8 & 36.8 & 23.2 & 37.6\\
    \hline
    Scans/sec & 0.1 & 30.0 & 0.04 & 12.9 & 0.03 & 12.2 \\
    \hline
    \end{tabular}}
    \vspace{-2mm}
\end{table*}

\begin{table}[!tb]
    \centering
    \setlength{\tabcolsep}{3.2mm}
    \caption{Ablation study for UnPNet.}
    \label{tab:ablation_unpnet}
    \vspace{-2mm}
    \resizebox{\linewidth}{!}{%
    \begin{tabular}{c|c|c|c}
    \hline
    & Acc & mIoU & Scans/s\\
    \hline
    RPointConv & 82.5 & 37.6 & 12.2\\
    \hline
    UnPNet $\textit{w/o}$ supp & 87.0 & 50.2 & 12.8\\
    UnPNet & 87.4 & 50.7 & 12.8\\
    UnPNet + $\textit{k}$-NN & 89.1 & 54.6 & 12.5\\
    \hline
    \end{tabular}}
\end{table}

\subsection{Auxiliary Supervision}
\label{sec:supervision}

For the decoder of UnPNet, we employ edge supervision to obtain accurate boundaries of the segments. 
Before each upsampling operation, we apply edge supervision by computing the edge loss $\mathcal{L}_{e}$ using the binary entropy loss as 
\begin{equation} \label{eq:edge}
    \mathcal{L}_e = -\frac{1}{|I|}\sum_{i\in\mathbf{I}}(e_i \log(\hat{e}_i) + (1-e_i)\log(1-\hat{e}_i)),
\end{equation}
where $e_i \in\{0,1\}$ is the ground-truth edge for pixel $i$ and $\hat{e}_i$ is the corresponding predicted probability estimated by a $1\times 1$ convolution. 
The ground-truth edge is obtained by the segment boundaries of the ground-truth segmentation.

As for the final semantic prediction of UnPNet, we use two loss terms.
The first one is the standard weighted cross-entropy loss which can be formulated as
\begin{equation}\label{eq:S}
    \mathcal{L}_s = -\frac{1}{|I|}\sum_{i\in I}\sum_{n=1}^{N}w_n p_i^n \log(\hat{p}_i^n),
\end{equation}
where $N$ is the number of classes, $p_i^n \in \{0,1\}$ is 1 if pixel $i$ is annotated by class $n$, and $\hat{p}_i^n$ is the predicted class probability. 
The weight $w_n$ for class $n$ is inversely proportional to its frequency of occurrence.

Apart from the weighted cross-entropy loss, we also directly maximize the intersection-over-union (IoU) score by the Lov\'asz-Softmax loss \cite{berman2018lovasz}:
\begin{equation}\label{eq:ls}
\mathcal{L}_{l s}=\frac{1}{N} \sum_{n=1}^{N} \overline{\Delta_{J_{n}}}(m(n)),
\end{equation}
\begin{equation}
m_{i}(n)=\left\{\begin{array}{ll}
1 - \hat{p}_i^n & \text{if } p_i^n=1 \\
\hat{p}_i^n & \text{otherwise},
\end{array}\right.
\end{equation}
where $\overline{\Delta_{J_{n}}}$ is the Lov\'asz extension of the Jaccard index. 

Hence, the total loss is given by
\begin{equation}
    \mathcal{L} = \mathcal{L}_\textit{\text{s}} + \mathcal{L}_\textit{\text{ls}} + \frac{1}{2}\sum_u \mathcal{L}^u_{e},
\label{equ:loss}
\end{equation}
where $\mathcal{L}_\textit{\text{s}}$ denotes the weighted cross-entropy loss \equref{eq:S}, $\mathcal{L}_\textit{\text{ls}}$ denotes the Lov\'asz-Softmax loss \equref{eq:ls}, and $\mathcal{L}^u_{e}$ is the edge loss \equref{eq:edge} for each upsampling step $u$ of the decoder. 

\section{Experiments} \label{sec:exp}
\subsection{Experimental Setup}

We use four challenging datasets to evaluate our method:

\begin{figure}[t]
    \centering
    \includegraphics[trim={0 10 0 30},clip,width=\linewidth]{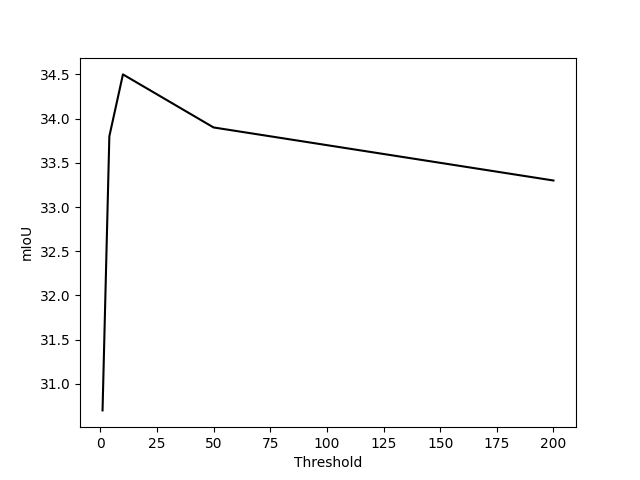}
    \vspace{-5mm}
    \caption{Effect of the radius.}
    \label{fig:threshold}
    \vspace{-3mm}
\end{figure}

\begin{table*}[t]
    \centering
    \caption{Evaluation results on the SemanticKITTI dataset.}
    \vspace{-2mm}
    \label{tab:semantickitti}
    \resizebox{\linewidth}{!}{%
    \begin{tabular}{c|ccccccccccccccccccc|c}
    \hline
          & 
         \rotatebox{90}{car} & 
         \rotatebox{90}{bicycle} & 
         \rotatebox{90}{motorcycle} & 
         \rotatebox{90}{truck} & 
         \rotatebox{90}{other-vehicle} & \rotatebox{90}{person} & 
         \rotatebox{90}{bicyclist} & 
         \rotatebox{90}{motorcyclist} & 
         \rotatebox{90}{road} & 
         \rotatebox{90}{parking} & 
         \rotatebox{90}{sidewalk} & 
         \rotatebox{90}{other-ground} &
         \rotatebox{90}{building} & 
         \rotatebox{90}{fence} & 
         \rotatebox{90}{vegetation} & 
         \rotatebox{90}{trunk} & 
         \rotatebox{90}{terrain} & 
         \rotatebox{90}{pole} & 
         \rotatebox{90}{traffic-sign} &
         \rotatebox{90}{mIoU} 
         \\ 
         \hline
         Pointnet \cite{qi2017pointnet} & 46.3 & 1.3 & 0.3 & 0.1 & 0.8 & 0.2 & 0.2 & 0.0 & 61.6 & 15.8 & 35.7 & 1.4 & 41.4 & 12.9 & 31.0 & 4.6 & 17.6 & 2.4 & 3.7 & 14.6 \\ 
         Pointnet++ \cite{qi2017pointnet++} & 53.7 & 1.9 & 0.2 & 0.9 & 0.2 & 0.9 & 1.0 & 0.0 & 72.0 & 18.7 & 41.8 & 5.6 & 62.3 & 16.9 & 46.5 & 13.8 & 30.0 & 6.0 & 8.9 & 20.1 \\
         SPGraph \cite{landrieu2018large} & 68.3 & 0.9 & 4.5 & 0.9 & 0.8 & 1.0 & 6.0 & 0.0 & 49.5 & 1.7 & 24.2 & 0.3 & 68.2 & 22.5 & 59.2 & 27.2 & 17.0 & 18.3 & 10.5 & 20.0 \\
         SPLATNet \cite{su2018splatnet} & 66.6 & 0.0 & 0.0 & 0.0 & 0.0 & 0.0 & 0.0 & 0.0 & 70.4 & 0.8 & 41.5 & 0.0 & 68.7 & 27.8 & 72.3 & 35.9 & 35.8 & 13.8 & 0.0 & 22.8 \\
         TangentConv \cite{tatarchenko2018tangent} & 86.8 & 1.3 & 12.7 & 11.6 & 10.2 & 17.1 & 20.2 & 0.5 & 82.9 & 15.2 & 61.7 & 9.0 & 82.8 & 44.2 & 75.5 & 42.5 & 55.5 & 30.2 & 22.2 & 35.9 \\
        \hline
         DeepLabV3+\cite{chen2018encoder} & 78.4 & 13.6 & 9.5 & 9.5 & 10.4 & 17.5 & 22.0 & 0.4 & 88.5 & 54.5 & 66.7 & 9.7 & 77.9 & 39.1 & 72.0 & 39.9 & 60.0 & 23.4 & 36.1 & 38.4 \\
         PSPNet\cite{li2018pyramid} & 79.6 & 25.0 & 26.4 & 17.5 & 24.0 & 34.1 & 28.4 & 7.3 & 90.2 & 58.2 & 70.2 & 19.9 & 79.7 & 43.5 & 74.2 & 43.2 & 61.2 & 23.1 & 37.5 & 44.4\\
         DenseASPP\cite{yang2018denseaspp} & 78.1 & 20.5 & 18.2 & 20.0 & 16.6 & 27.8 & 28.9 & 5.7 & 88.5 & 53.3 & 67.5 & 9.3 & 76.3 & 39.6 & 70.0 & 36.8 & 57.7 & 15.9 & 32.4 & 40.2\\
         \hline
         SqueezeSeg \cite{wu2018squeezeseg} & 68.8 & 16.0 & 4.1 & 3.3 & 3.6 & 12.9 & 13.1 & 0.9 & 85.4 & 26.9 & 54.3 & 4.5 & 57.4 & 29.0 & 60.0 & 24.3 & 53.7 & 17.5 & 24.5 & 29.5\\
         SqueezeSeg + CRF \cite{wu2018squeezeseg} & 68.3 & 18.1 & 5.1 & 4.1 & 4.8 & 16.5 & 17.3 & 1.2 & 84.9 & 28.4 & 54.7 & 4.6 & 61.5 & 29.2 & 59.6 & 25.5 & 54.7 & 11.2 & 36.3 & 30.8 \\
         SqueezeSegV2 \cite{wu2019squeezesegv2} & 81.8 & 18.5 & 17.9 & 13.4 & 14.0 & 20.1 & 25.1 & 3.9 & 88.6 & 45.8 & 67.6 & 17.7 & 73.7 & 41.1 & 71.8 & 35.8 & 60.2 & 20.2 & 36.3 &  39.7\\
         SqueezeSegV2 + CRF \cite{wu2019squeezesegv2} & 82.7 & 21.0 & 22.6 & 14.5 & 15.9 & 20.2 & 24.3 & 2.9 & 88.5 & 42.4 & 65.5 & 18.7 & 73.8 & 41.0 & 68.5 & 36.9 & 58.9 & 12.9 & 41.0 & 39.6 \\
         RangeNet21 \cite{milioto2019rangenet++} & 85.4 & 26.2 & 26.5 & 18.6 & 15.6 & 31.8 & 33.6 & 4.0 & 91.4 & 57.0 & 74.0 & 26.4 & 81.9 & 52.3 & 77.6 & 48.4 & 63.6 & 36.0 & 50.0 & 47.4 \\
         RangeNet53 \cite{milioto2019rangenet++} & 86.4 & 24.5 & 32.7 & \textbf{25.5} & 22.6 & 36.2 & 33.6 & 4.7 & \textbf{91.8} & \textbf{64.8} & \textbf{74.6} & \textbf{27.9} & 84.1 & \textbf{55.0} & 78.3 & 50.1 & 64.0 & 38.9 & 52.2 & 49.9\\
         \hline
         RPointNet++ & 73.3 & 13.0 & 5.4 & 11.8 & 8.3 & 6.4 & 15.5 & 2.1 & 86.3 & 40.1 & 60.1 & 7.2 & 61.7 & 32.1 & 55.8 & 13.1 & 51.6 & 4.2 & 14.7 & 29.6\\
        RSCNN & 79.8 & 19.8 & 11.2 & 15.8 & 14.3 & 15.1 & 20.5 & 8.8 & 87.1 & 41.8 & 64.7 & 8.7 & 72.2 & 37.9 & 68.0 & 28.4 & 58.0 & 13.1 & 31.3 & 36.7\\
        RPointConv & 79.5 & 19.0 & 12.7 & 13.8 & 10.7 & 14.9 & 18.2 & 5.8 & 87.8 & 46.6 & 66.6 & 7.3 & 73.2 & 40.1 & 69.4 & 30.9 & 59.3 & 14.1 & 32.1 & 36.9 \\
         UnPNet & 70.8 & 38.2 & 31.8 & 20.3 & 22.5 & 49.0 & 45.8 & 14.3 & 91.4 & 61.6 & 73.6 & 19.2 & 82.4 & 50.8 & 77.7 & 51.6 & 65.3 & 34.9 & 53.6 & 51.0 \\
         UnPNet + $\textit{k}$-NN& \textbf{90.4} & \textbf{43.8} & \textbf{36.1} & 20.4 & \textbf{23.1} & \textbf{54.3} & \textbf{54.2} & \textbf{14.4} & 91.4 & 62.0 & 74.2 & 18.9 & \textbf{86.3} & 54.6 & \textbf{80.5} & \textbf{59.4} & \textbf{66.3} & \textbf{48.7} & \textbf{58.6} & \textbf{54.6} \\
         \hline
    \end{tabular}}
\end{table*}

\begin{table*}[!t]
    \centering
    \caption{Evaluation results on the nuScenes dataset.}
    \vspace{-2mm}
    \label{tab:nuscenes}
    \resizebox{\linewidth}{!}{%
    \begin{tabular}{c|cccccccccccccccc|c}
    \hline
           & 
         \rotatebox{90}{barrier} & 
         \rotatebox{90}{bicycle} & 
         \rotatebox{90}{bus} & 
         \rotatebox{90}{car} & 
         \rotatebox{90}{cons\_vehicle} & 
         \rotatebox{90}{motorcycle} & 
         \rotatebox{90}{pedestrian} & 
         \rotatebox{90}{traffic\_cone} & 
         \rotatebox{90}{trailer} & 
         \rotatebox{90}{truck} & 
         \rotatebox{90}{driv\_surf} & 
         \rotatebox{90}{other\_flat} &
         \rotatebox{90}{sidewalk} &
         \rotatebox{90}{terrain} &
         \rotatebox{90}{manmade} &
         \rotatebox{90}{vegetation} &
         \rotatebox{90}{mIoU} \\
         \hline
         DeepLabV3\cite{chen2018encoder} & 50.5 & 4.9 & 58.6 & 60.3 & 10.6 & 7.5 & 21.6 & 17.2 & 35.1 & 46.2 & 88.1 & 47.2 & 55.7 & 59.2 & 69.1 & 69.4 & 43.8\\
         DenseASPP\cite{yang2018denseaspp} & 52.7 & 6.2 & 69.0 & 67.1 & 18.5 & 32.6 & 24.3 & 16.9 & 39.7 & 58.0 & 87.4 & 43.1 & 57.1 & 58.6 & 69.8 & 70.4 & 48.2 \\
         PSPNet\cite{zhao2017pyramid} & 56.9 & 8.9 & 69.8 & 68.9 & \textbf{23.6} & 36.9 & 26.5 & 18.9 & \textbf{42.3} & 58.7 & 88.0 & 44.5 & 58.5 & 59.6 & 71.5 & 71.6 & 50.3 \\ 
         \hline
        SqueezeSegV1\cite{wu2018squeezeseg} &  15.0 & 0.3 & 4.5 & 25.8 & 0.0 & 0.5 & 3.1 & 5.3 & 2.5 & 9.4 & 68.3 & 11.2 & 23.3 & 42.1 & 45.6 & 40.1 & 18.6\\
         SqueezeSegV2\cite{wu2019squeezesegv2} &  44.3 & 2.7 & 62.2 & 68.0 & 11.2 & 19.3 & 7.6 & 12.1 & 25.3 & 44.8 & 84.8 & 29.8 & 51.6 & 56.6 & 64.4 & 67.8 & 40.8 \\ 
         RangeNet21\cite{milioto2019rangenet++} &  61.4 & 3.4 & 72.9 & 76.4 & 17.2 & 23.5 & 31.5 & 19.3 & 35.8 & 59.8 & 92.3 & 54.6 & 66.5 & 68.1 & 76.9 & 76.6 & 52.3\\ 
         RangeNet53\cite{milioto2019rangenet++} & 59.8 & 2.7 & 62.6 & 73.5 & 14.1 & 21.6 & 28.3 & 13.9 & 34.5 & 58.3 & 90.8 & 53.8 & 61.7 & 64.5 & 74.8 & 75.0 & 49.4 \\ 
         \hline
         UnPNet & \textbf{61.2} & 5.9 & \textbf{77.7} & 73.2 & 21.9 & 34.7 & 38.5 & 25.7 & 40.3 & 62.9 & 92.3 & 61.6 & 66.7 & 67.7 & 78.7 & 78.9 & 55.5\\ 
         UnPNet + $\textit{k}$-NN & 61.0 & \textbf{6.3} & \textbf{77.7} & \textbf{78.4} & 21.9 & \textbf{37.0} & {42.5} & \textbf{30.5} & 41.7 & \textbf{65.8} & \textbf{93.8} & \textbf{62.2} & \textbf{66.9} & \textbf{68.1} & \textbf{80.6} & \textbf{80.0} & \textbf{57.2} \\
         \hline
    \end{tabular}}
\end{table*}

\begin{itemize}
    \item \textbf{SemanticKITTI} \cite{behley2019semantickitti} provides pixel-wise semantic labels for the entire KITTI Odometry Benchmark~\cite{geiger2012we} with more than 20000 scans which are divided into 22 sequences.
    We use its training set for training (sequences 00-10 without sequence 08), its validation set (sequence 08) for the ablation study, and its test set (sequences 10-21) for the comparison with state-of-the-art methods.
    \item \textbf{SemanticPOSS} \cite{pan2020semanticposs} contains about 3000 scans at the Peking University, which are divided into 6 equal subsets.
    For the experiments, we use the 3rd subset for evaluation and the others for training as in \cite{pan2020semanticposs}.
    There are usually more moving and small objects in the campus scenarios, making it more difficult compared to urban environments.
    \item \textbf{nuScenes} \cite{caesar2020nuscenes} includes about 40000 annotated scans captured in 900 scenes, where 750 scenes are for training and the others for validation.
    As recommended\footnote{The nuScenes LiDAR segmentation task is available at \href{https://www.nuscenes.org/lidar-segmentation}{https://www.nuscenes.org/lidar-segmentation}.}, we merge similar classes and remove rare classes.
    \item \textbf{Pandaset}\footnote{\href{https://scale.com/open-datasets/pandaset}{https://scale.com/open-datasets/pandaset}.} 
    contains about 16000 LiDAR scans at 2 routes in the Silicon Valley.
    Two LiDAR sensors have been used for recording the data, a spinning LiDAR and a solid-state LiDAR.
    For the experiments, the data from the spinning LiDAR is used.
    We use 30\% of the data for evaluation and the rest for training.
    Similar to the nuScenes dataset, we merge similar classes and remove rare classes.
\end{itemize}
%
As for the evaluation metric, we use the standard mean intersection over union (mIoU) metric~\cite{everingham2015pascal} over all classes.
For a fair comparison, all experiments are performed with a single GPU.

\subsection{Ablation Study}
The ablation study is conducted on the SemanticKITTI dataset.

\subsubsection{Effect of hyperparameters}
We first explore the influence of the size $k$ of the 2D search region in the reformulated grouping layer. For the study, we use the reformulated PointNet++ (RPointNet++) with an input resolution of $64\times 512$.
The results are shown in \tabref{tab:ablation}~(a).
We can see that the performance improves and the speed slows down with the search size $k$ increasing.
However, the improvement from $k=5$ to $k=7$ is not as large as that from $k=3$ to $k=5$, which indicates that the local region with $k=5$ already includes the most important neighboring points.
Considering the trade-off between effectiveness and efficiency, we set $k=5$ in the following experiments.

Then, we evaluate the effect of different input resolutions, and we summarize the results in \tabref{tab:ablation}~(b).
We can observe that increasing the input resolution from 64$\times$512 to 64$\times$1024 improves the accuracy but at the cost of higher inference time. However, when the resolution changes from 64$\times$1024 to 64$\times$2048, the accuracy degrades. Since we kept $k$ constant, increasing the resolution decreases the receptive field. Hence, $k$ needs to be increased when the resolution increases.      
For a good trade-off between effectiveness and efficiency, we set the input resolution to 64$\times$512 if not mentioned otherwise.

We also evaluate the impact of the stride of the uniform sampling, i.e., $H' = \frac{H}{S}$ and $W' = \frac{W}{S}$. 
The results are shown in \tabref{tab:ablation}~(c).
The accuracy is similar for the sampling stride 1 and 2, but using stride 2 is more efficient.
When the sampling stride is set to 4, the accuracy drops significantly since the sampling becomes too sparse. We use therefore sampling stride 2. 

\begin{table*}[!t]
    \centering
    \caption{Evaluation results on the Pandaset dataset.}
    \label{tab:pandaset}
    \vspace{-2mm}
    \resizebox{\linewidth}{!}{%
    \begin{tabular}{c|ccccccccccccccccc|c}
    \hline
           & 
         \rotatebox{90}{Vegetation} & 
         \rotatebox{90}{Ground} & 
         \rotatebox{90}{Road} & 
         \rotatebox{90}{Lane Line } & 
         \rotatebox{90}{Road Mark} & 
         \rotatebox{90}{Sidewalk} & 
         \rotatebox{90}{Car} & 
         \rotatebox{90}{Truck} & 
         \rotatebox{90}{Motorcycle} & 
         \rotatebox{90}{Bus} &
         \rotatebox{90}{Bicycle} &
         \rotatebox{90}{Pedestrian} &
         \rotatebox{90}{Pylons} &
         \rotatebox{90}{Signs} &
         \rotatebox{90}{Cones} &
         \rotatebox{90}{Const-Signs} &
         \rotatebox{90}{Building} &
         \rotatebox{90}{mIoU} \\
         \hline
         DeepLabV3\cite{chen2018encoder} & 50.4 & 22.8 & 69.2 & 11.9 & 9.7 & 35.5 & 65.6 & 11.6 & 0.1 & 3.9 & 0.2 & 3.5 & 2.2 & 19.2 & 4.2 & 11.5 & 58.4 & 22.3 \\
         DenseASPP\cite{yang2018denseaspp} & 54.6 & 21.2 & 66.0 & 9.9 & 9.3 & 33.2 & 64.9 & 18.8 & 0.0 & 17.6 & 0.2 & 5.0 & 0.3 & 21.2 & 2.7 & 4.6 & 62.5 & 23.1 \\
         PSPNet\cite{zhao2017pyramid} & 45.9 & 18.1 & 65.4 & 8.4 & 6.8 & 28.1 & 60.2 & 15.6 & 0.0 & 6.2 & 0.2 & 2.5 & 0.0 & 19.0 & 0.9 & 4.5 & 55.2 & 19.8 \\
         \hline
        SqueezeSegV1\cite{wu2018squeezeseg} & 40.0 & 16.5 & 66.7 & 9.4 & 6.3 & 19.2 & 52.4 & 9.6 & 0.0 & 3.9 & 0.3 & 2.6 & 0.1 & 15.5 & 1.2 & 0.0 & 39.4 & 16.6 \\ 
         SqueezeSegV2\cite{wu2019squeezesegv2} & 61.0 & 33.0 & 76.4 & 18.6 & 13.5 & 43.5 & 73.5 & 30.8 & 0.0 & 21.8 & 1.7 & 4.9 & 1.1 & 20.5 & 3.3 & 3.7 & 65.4 & 27.8 \\ 
         RangeNet21\cite{milioto2019rangenet++} & 65.9 & 35.3 & 81.1 & 26.6 & 20.0 & 47.1 & 75.0 & 28.6 & 0.1 & \textbf{25.7} & 1.5 & 12.0 & 1.6 & 34.2 & 4.9 & 19.0 & 72.4 & 32.4 \\  %
         RangeNet53\cite{milioto2019rangenet++} & 64.1 & 36.3 & 80.8 & 25.1 & 19.8 & 48.8 & 76.0 & 30.7 &\textbf{1.0} & 17.2 & 2.0 & 9.8 & 6.3 & 32.7 & 6.7 & 13.0 & 71.0 & 31.8\\ %
         \hline
         UnPNet & 75.4 & 40.7 & 82.4 & 26.6 & 19.9 & 49.8 & 76.4 & 31.1 & 0.7 & 17.5 & 10.4 & 29.5 & 28.6 & 44.4 & 14.5 & 19.5 & 79.6 & 38.1 \\%
         UnPNet + $\textit{k}$-NN & \textbf{78.2} & \textbf{41.6} & \textbf{82.5} & \textbf{29.3} & \textbf{21.7} & \textbf{52.0} & \textbf{78.9} & \textbf{31.9} & 0.9 & 17.7 & \textbf{12.1} & \textbf{36.7} & \textbf{40.4} & \textbf{63.7} & \textbf{19.5} & \textbf{24.9} & \textbf{83.8} & \textbf{42.1} \\ 
         \hline
    \end{tabular}}
    \vspace{-2mm}
\end{table*}

\begin{table*}[!t]
    \centering
    \caption{Evaluation results on the SemanticPOSS dataset. }
    \vspace{-2mm}
    \label{tab:semanticposs}
    \resizebox{\linewidth}{!}{%
    \begin{tabular}{c|ccccccccccc|c}
    \hline
           & 
         {person} & 
         {rider} & 
         {car} & 
         {trunk} & 
         {plants} & 
         {traffic sign} & 
         {pole} & 
         {building} & 
         {fence} & 
         {bike} &
         {road} &
         {mIoU} \\
         \hline
         DeepLabV3\cite{chen2018encoder} &  9.6 & 4.8 & 15.3 & 9.5 & 40.3 & 5.3 & 3.1 & 41.2 & 11.9 & 20.0 & 62.5&20.3 \\
         DenseASPP\cite{yang2018denseaspp} & 11.6&5.7&20.4&10.8&46.4&5.2&4.9&46.8&7.7&20.6&62.0&22.0\\
         PSPNet\cite{zhao2017pyramid} & 9.0&4.8&17.6&10.3&44.1&5.1&3.2&45.3&5.9&20.0&63.0&20.8\\
         \hline
         SqueezeSegV1\cite{wu2018squeezeseg} & 5.5&0.0&8.7&3.4&39.1&2.4&2.5&34.5&7.6&18.4&62.5&16.8 \\ 
         SqueezeSegV2\cite{wu2019squeezesegv2} & \textbf{18.4} & 11.2 & 34.9 & \textbf{15.8} & 56.3 & \textbf{11.0} & 4.5 & 47.0 & 25.5 & 32.4 & \textbf{71.3}  & 29.8\\
         RangeNet21\cite{milioto2019rangenet++} & 9.8&7.8&\textbf{39.9}&8.0&55.8&7.0&2.9&50.3&19.2&32.3&63.8&27.0 \\  
         RangeNet53\cite{milioto2019rangenet++} & 10.0&6.2&33.4&7.3&54.2&5.5&2.6&49.9&18.4&28.6&63.5&25.4 \\ 
         \hline
         UnPNet & 11.3&12.1&36.8&10.6&62.3&6.9&4.2&60.4&20.6&35.4&65.6&29.7 \\ 
         UnPNet + $\textit{k}$-NN & 
         17.7&\textbf{17.2}&39.2&13.8&\textbf{67.0}&9.5&\textbf{5.8}&\textbf{66.9}&\textbf{31.1}&\textbf{40.5}&68.4&\textbf{34.3} \\
         \hline
    \end{tabular}}
    \vspace{-2mm}
\end{table*}

\begin{figure}[!t]
    \centering
    \includegraphics[width=\linewidth]{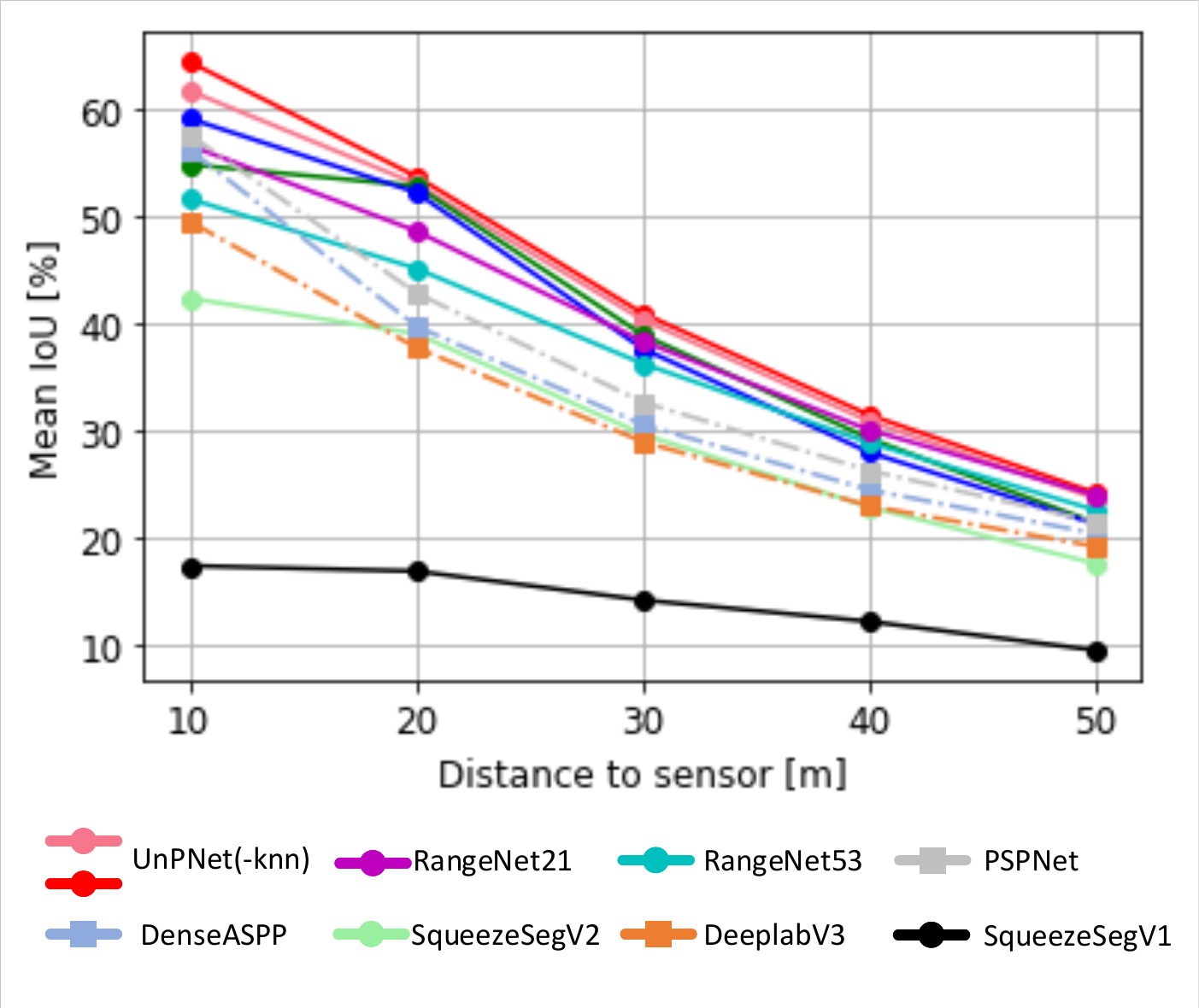}
    \caption{Mean IoU with regard to the distance of the points to the LiDAR sensor, evaluated on the nuScenes dataset \cite{caesar2020nuscenes}.
    The dashed lines denote image-based methods while solid lines represent projection-based methods. For UnPNet, we show both the results with $k$-NN (red) and without $k$-NN post-processing (light red).}
    \label{fig:miou_dist}
\end{figure}

\begin{figure*}[htbp]
\vspace{-1mm}
\centering
\subfloat[RPointNet++]{\includegraphics[width=0.44\linewidth,height=3cm]{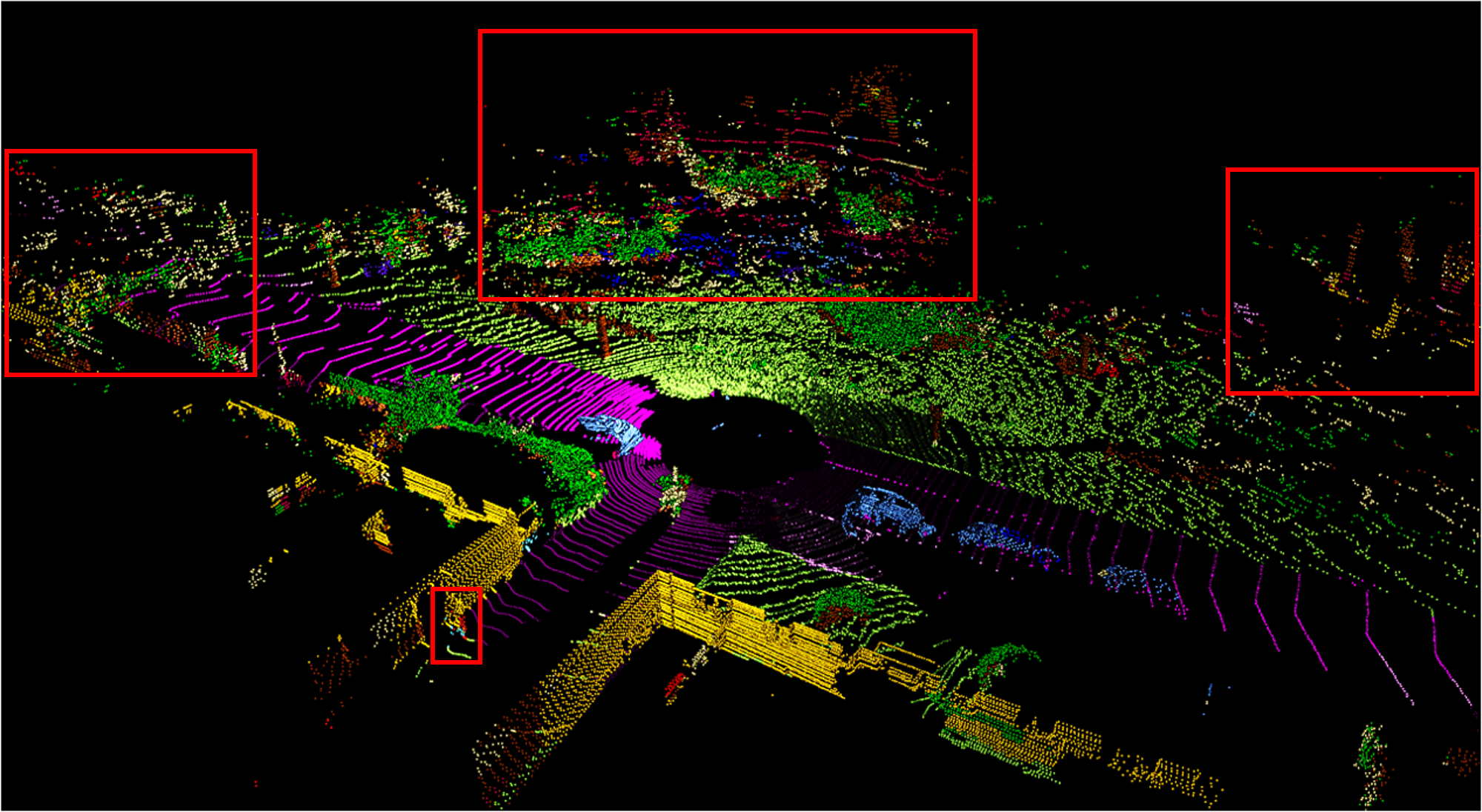}} \hspace{4mm}
\subfloat[RSCNN]{\includegraphics[width=0.44\linewidth,height=3cm]{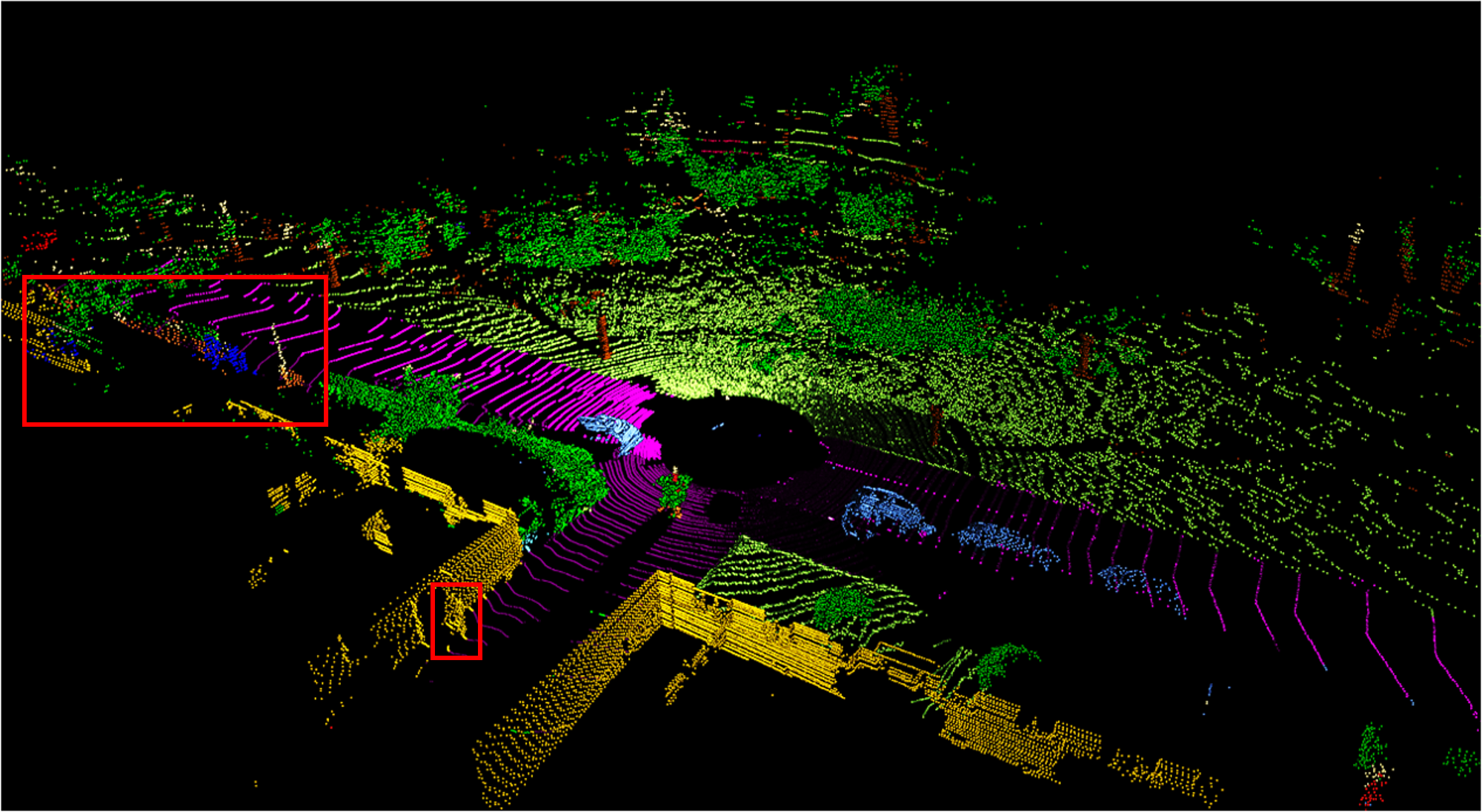}} \
\ 
\subfloat[RPointConv]{\includegraphics[width=0.44\linewidth,height=3cm]{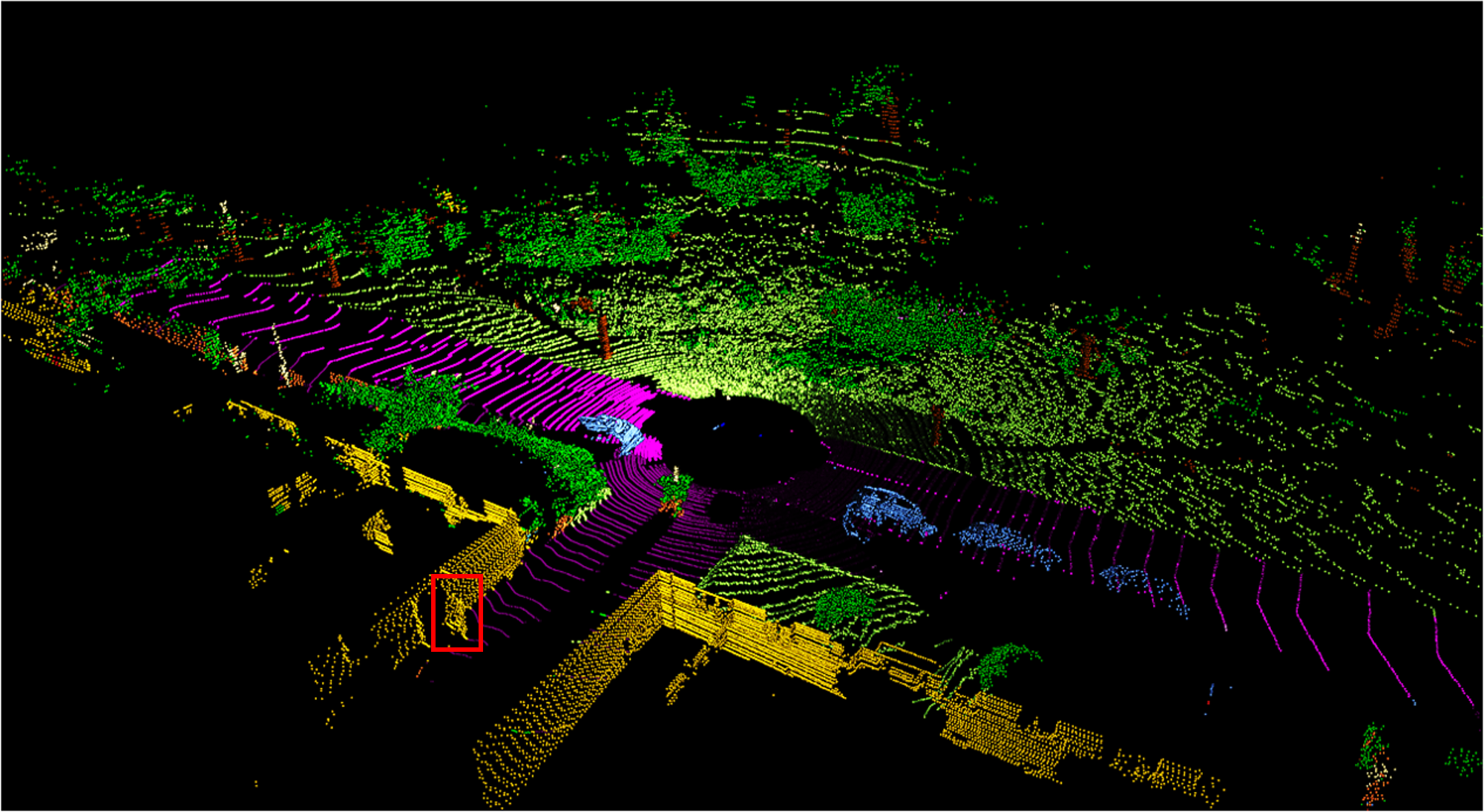}} \hspace{4mm}
\subfloat[Ground Truth]{\includegraphics[width=0.44\linewidth,height=3cm]{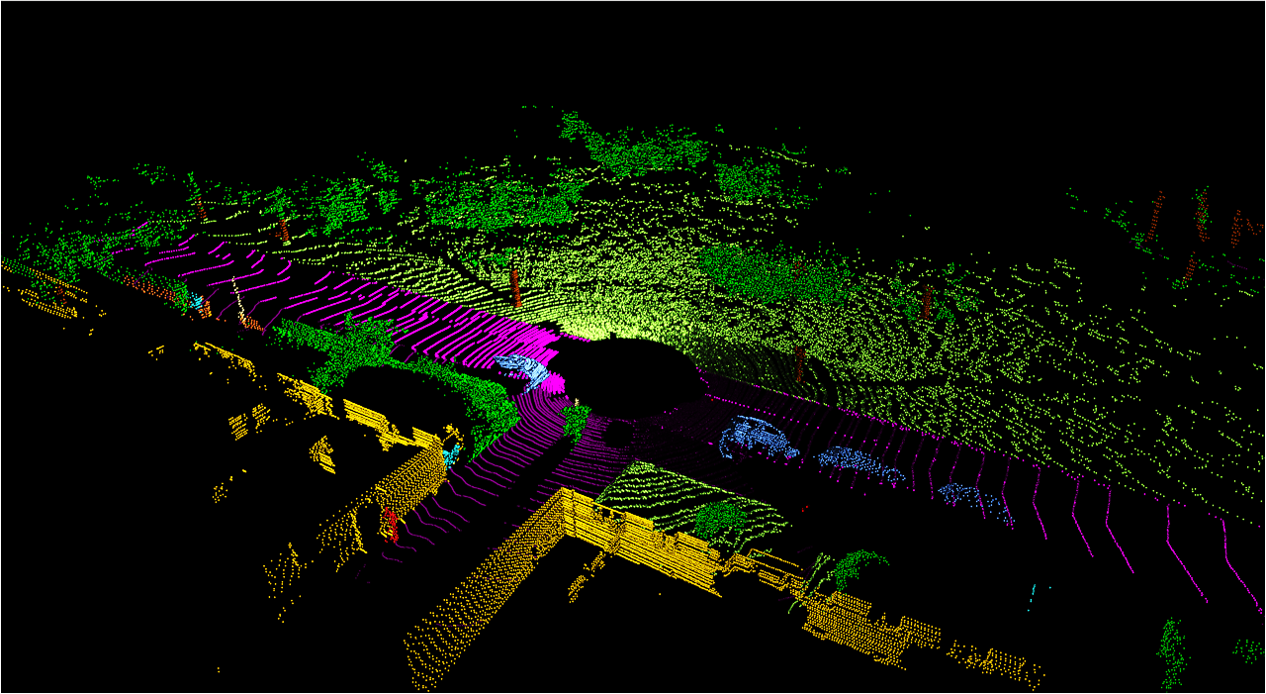}}
\caption{Qualitative results of RPointNet++, RSCNN, and RPointConv. We highlight wrong predictions by the red boxes. RPointNet++ makes wrong predictions for distant points due to the weak aggregation capability of the used pooling operations. RPointConv achieves better results than RSCNN, demonstrating the effectiveness of density information.}
\label{fig:visualization}
\vspace{-3mm}
\end{figure*}

As for the  impact of the radius, we vary it between 1m and 200m. The results are shown in \figref{fig:threshold}. The mIoU increases until a radius of 10m and then slightly decreases. We use 10m as the default threshold. 

\begin{figure*}[!t]
\vspace{-1mm}
\centering
\subfloat[Ground Truth]{
\begin{minipage}[t]{0.49\linewidth}
\centering
\includegraphics[trim={150 20 0 30},clip,width=\linewidth,height=3.8cm]{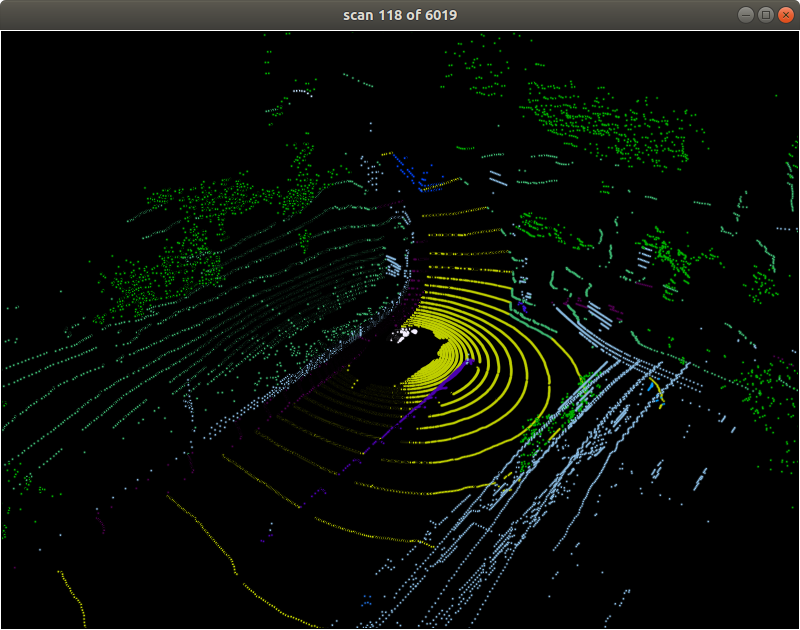}\\ \vspace{2mm}
\includegraphics[trim={150 20 0 30},clip,width=\linewidth,height=3.8cm]{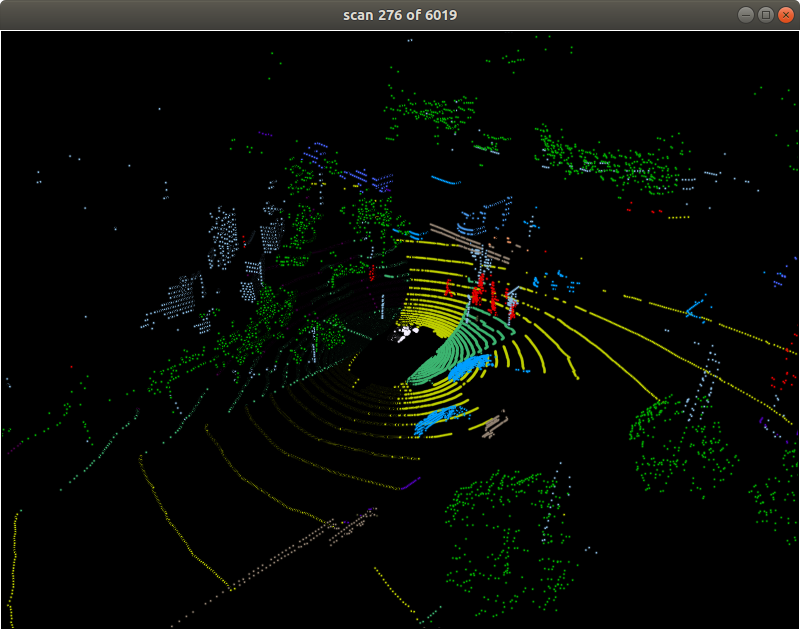}\\
\vspace{2mm}
\includegraphics[trim={150 20 0 30},clip,width=\linewidth,height=3.8cm]{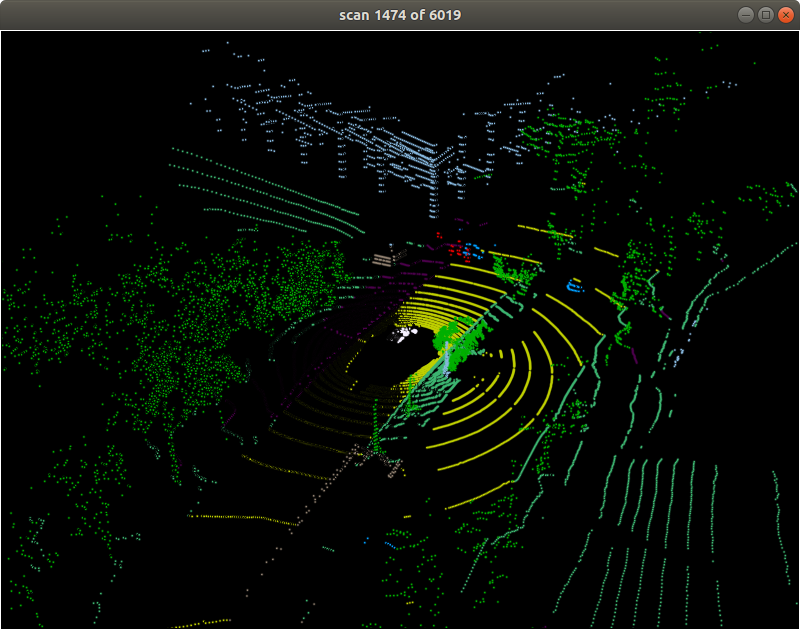}\\
\end{minipage}
} 
\subfloat[UnPNet]{
\begin{minipage}[t]{0.49\linewidth}
\centering
\includegraphics[trim={150 20 0 30},clip,width=\linewidth,height=3.8cm]{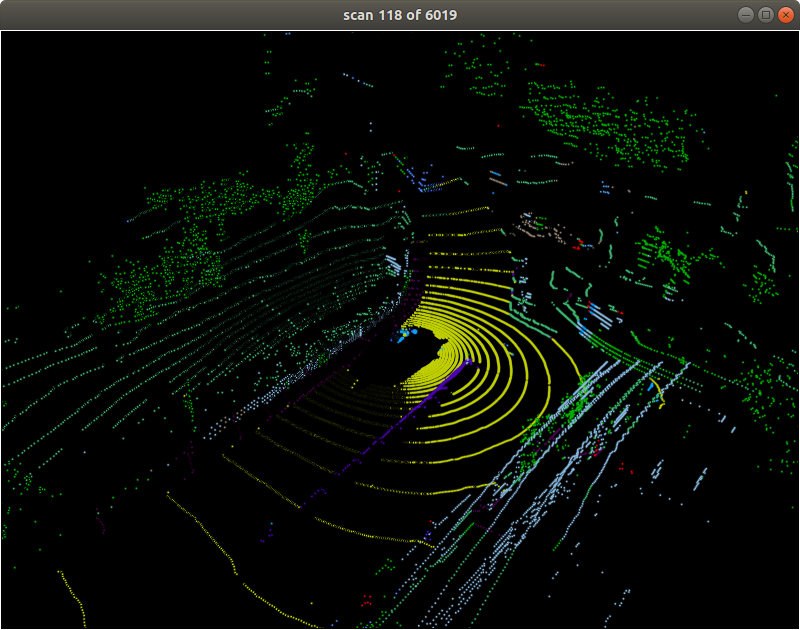}\\ \vspace{2mm}
\includegraphics[trim={150 20 0 30},clip,width=\linewidth,height=3.8cm]{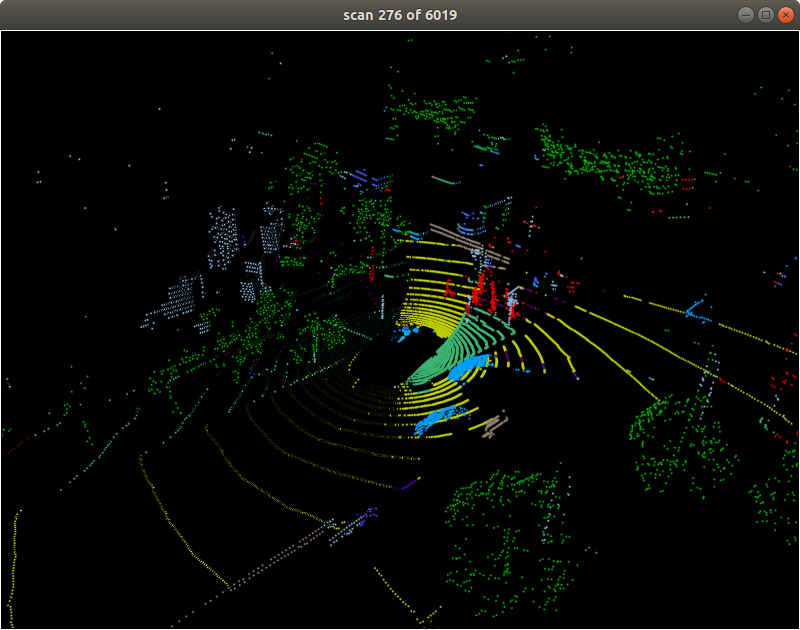}\\
\vspace{2mm}
\includegraphics[trim={150 20 0 30},clip,width=\linewidth,height=3.8cm]{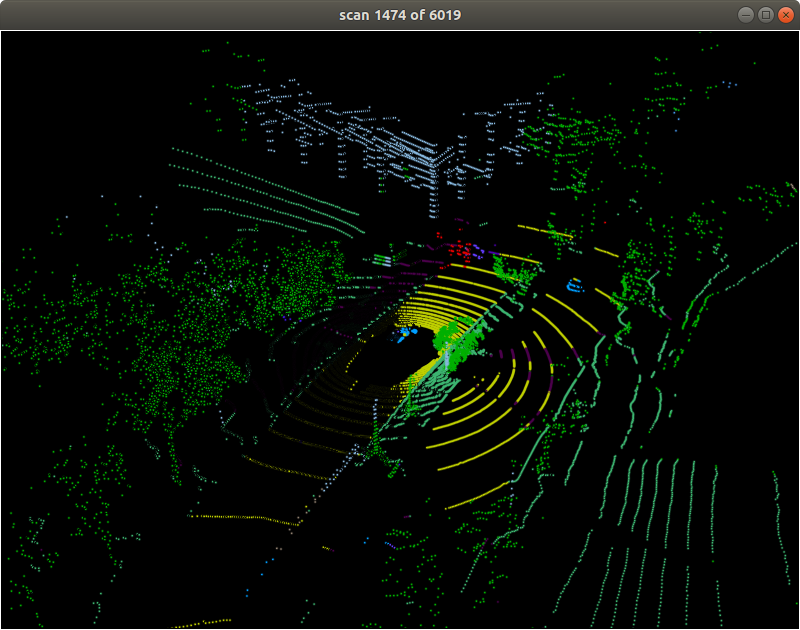}\\
\end{minipage}
} 
\caption{Qualitative results of UnPNet for the nuScenes dataset. 
}
\label{fig:vvis}
\vspace{-3mm}
\end{figure*}

\subsubsection{Comparison between reformulated and original methods}

Finally, we compare the original models with their reformulated versions.
The three baseline methods are PointNet++~\cite{qi2017pointnet++}, SpiderCNN~\cite{xu2018spidercnn} and PointConv~\cite{wu2019pointconv}.
We display the results in \tabref{tab:comparision_origin}.
We can see that the reformulated point-based methods achieve a much better performance than the original baselines.
The improvement is about 10\% for all baselines in terms of both accuracy and mIoU.
For SpiderCNN~\cite{xu2018spidercnn} and PointConv~\cite{wu2019pointconv}, the mIoU is even improved by about 15\%.
These results prove the effectiveness of the reformulated point-based methods.
Apart from the mIoU and accuracy, the efficiency is also much better.
We can see that the reformulated methods are more than $300\times$ faster compared with the original ones.
This is because the reformulated methods utilize the projection space such that the 3D operations can be performed much more efficiently. This demonstrates how point-based methods can be improved by reformulating them. \figref{fig:visualization} shows some qualitative results of the three reformulated versions.

\subsubsection{Ablation study for UnPNet}
After having discussed the reformulated point-based networks, we now evaluate UnPNet and the results are shown in \tabref{tab:ablation_unpnet}. Although UnPNet uses the reformulated feature propagation from RPointConv, it outperforms RPointConv by a large margin. While the inference time is very similar, the mIoU is much higher for UnPNet compared to RPointConv. This demonstrates that leveraging reformulated 3D point-based operations and 2D image-based operations achieves a good performance. We also analyze the impact of the edge supervision. Without edge supervision, the mIoU decreases by 0.5\%. The $\textit{k}$-NN post-processing increases the accuracy. 

\subsection{Comparison with State-of-the-art}
We compare the proposed reformulated networks RPointNet++, RSCNN, and RPointConv as well as UnPNet with other methods. For a comprehensive comparison, we selected point-based methods (Pointnet \cite{qi2017pointnet}, Pointnet++ \cite{qi2017pointnet++}, SPGraph \cite{landrieu2018large}, SPLATNet \cite{su2018splatnet}, TangentConv \cite{tatarchenko2018tangent}), image-based methods (DeepLabV3 \cite{chen2018encoder}, DenseASPP \cite{yang2018denseaspp}, PSPNet \cite{zhao2017pyramid}) and projection-based methods (SqueezeSeg \cite{wu2018squeezeseg,wu2019squeezesegv2}, RangeNet \cite{milioto2019rangenet++}).

We first show the results for the SemanticKITTI test dataset in \tabref{tab:semantickitti}. While the reformulated point-based methods RPointNet++, RSCNN, and RPointConv outperform the corresponding baselines PointNet++, SCNN, and PointConv as on the validation set, they do not achieve the accuracy of state-of-the-art approaches. However, as we discussed, the general concept of reformulating point-based methods can also be used to develop new architectures by leveraging reformulated 3D operations and 2D CNNs. This is done by the proposed UnPNet and we can see that it outperforms RPointNet++, RSCNN, and RPointConv by a large margin. Using $\textit{k}$-NN as in \cite{milioto2019rangenet++} for post-processing improves the accuracy further. Compared to the other approaches, UnPNet performs in particular well for small objects like bicycle, person, or motorcyclist.


The results for the nuScenes dataset are shown in \tabref{tab:nuscenes}.
Similar to the SemanticKITTI dataset, UnPNet outperforms the other methods by a large margin. Furthermore, it achieves the best performance for almost all classes. In \figref{fig:miou_dist}, we also report the mIoU based on the distance to the sensor. We can see that as the points are more distant to the LiDAR sensor, the accuracy decreases as expected. Nevertheless, UnPNet outperforms the other methods for all distances. We can also observe that the $k$-NN post-processing mainly improves the accuracy at the close distances. In \figref{fig:vvis}, we show some qualitative results for UnPNet.  

For the Pandaset dataset, UnPNet can successfully detect some small objects that other methods hardly recognize like bicycle or cones which can be seen in \tabref{tab:pandaset}. The overall performance of our method is more than 10\% higher compared with the other methods.
Finally, we present the results for the SemanticPOSS dataset in \tabref{tab:semanticposs}. Our method also outperforms the other methods and improves mIoU by about 4\%. 
In summary, UnPNet achieves the best performance on all four datasets. This shows the robustness of UnPNet. 

\section{Conclusion} \label{sec:conclusion}
In this paper, we proposed a new paradigm to reformulate point-based methods so that they operate in the projection space of a LiDAR sensor. As examples, we used three point-based approaches and demonstrated that reformulated point-based methods achieve a better segmentation accuracy and efficiency than the original point-based methods. Furthermore, we proposed a new architecture named Unprojection Network (UnPNet) which combines the benefits of both point-based and projection-based methods. On one hand, it extracts features efficiently like projection-based methods and fuses the context information from different 2D scales. On the other hand, it exploits the full 3D structure as point-based methods for down- and upsampling. In addition, auxiliary edge supervision is utilized, which is infeasible for point-based methods. We evaluated the approach on four challenging datasets for semantic LiDAR point cloud segmentation and showed that the proposed concept of reformulating 3D point-based operations allows to design new architectures that outperform point- and projection-based approaches.
\add{As our approach of reformulating point-based operations is generic, it can be also used to make point-voxel methods like PV-RCNN \cite{shi2020pv} more efficient by reformulating the point-based operations within the network. We will explore this in the future work.}


%





\ifCLASSOPTIONcaptionsoff
  \newpage
\fi



%



{\small
\bibliographystyle{IEEEtran}
\bibliography{references}
}

%

\vspace{-0.5in}

\begin{IEEEbiography}[{\includegraphics[width=1in,height=1.25in,clip,keepaspectratio]{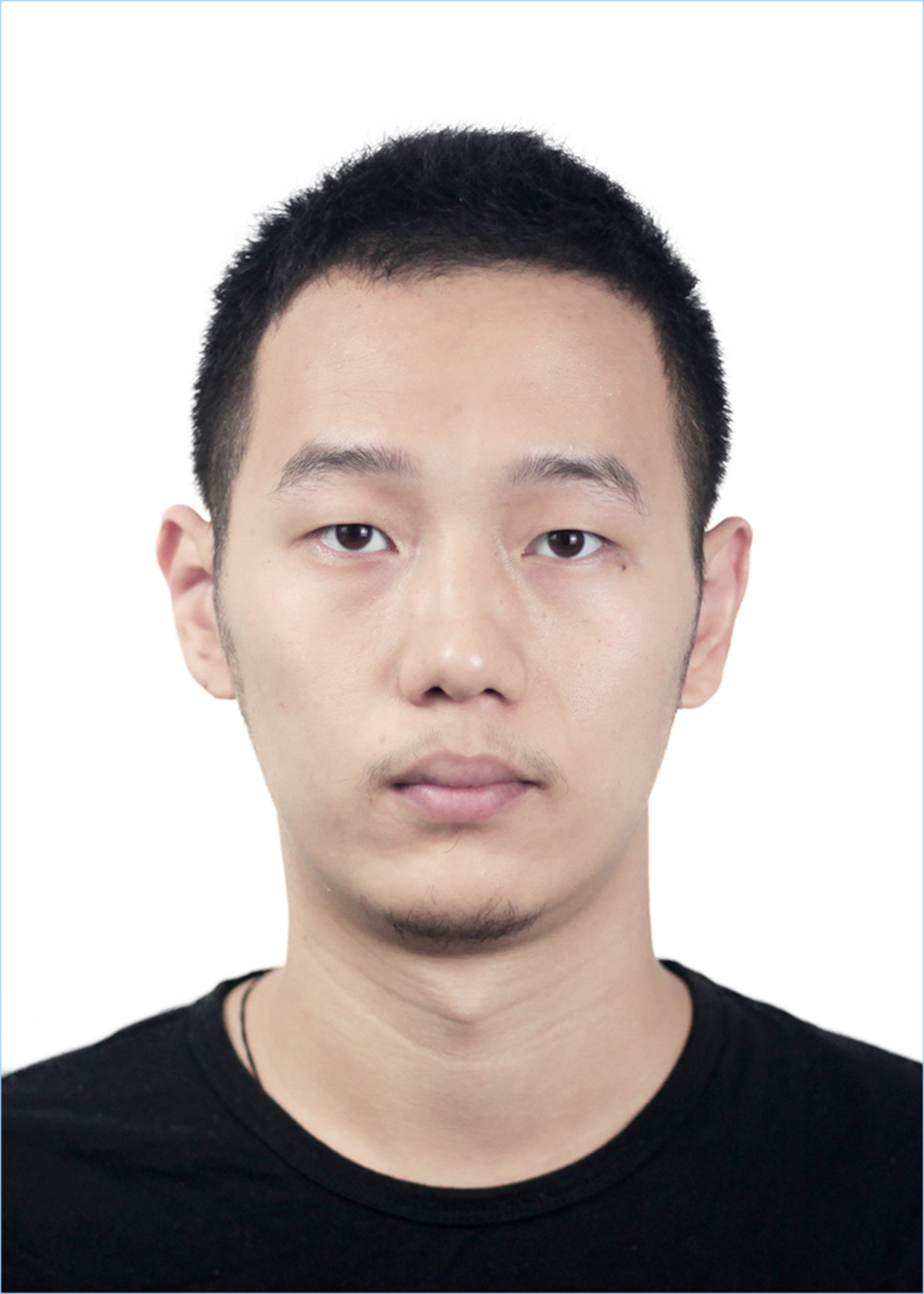}}]{Shijie Li}
received his bachelor's degree in Automation Engineering from University of Electronic Science and Technology of China in 2016 and his master's degree in computer science from the Nankai University in 2019. Since 2019, he is a Ph.D. student at the University of Bonn. His research interests include action recognition and scene understanding.
\end{IEEEbiography}

\vspace{-0.5in}

\begin{IEEEbiography}[{\includegraphics[width=1in,height=1.25in,clip,keepaspectratio]{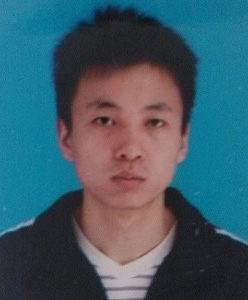}}]{Yun Liu}
received his bachelor's and Ph.D. degrees from Nankai University in 2016 and 2020, respectively. Since 2021, he is a postdoctoral researcher at the Computer Vision Laboratory, ETH Zurich.
His research interests include computer vision and machine learning.
\end{IEEEbiography}

\vspace{-0.5in}


\begin{IEEEbiography}[{\includegraphics[width=1in,height=1.25in,clip,keepaspectratio]{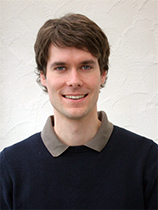}}]{Juergen Gall}
obtained his bachelor's and his master's degree in mathematics from the University of Wales Swansea (2004) and from the University of Mannheim (2005). 
In 2009, he obtained a Ph.D. in computer science from the Saarland University and the Max Planck Institut für Informatik. 
He was a postdoctoral researcher at the Computer Vision Laboratory, ETH Zurich, from 2009 until 2012 and senior research scientist at the Max Planck Institute for Intelligent Systems in Tübingen from 2012 until 2013. 
Since 2013, he is a professor at the University of Bonn and head of the Computer Vision Group.
\end{IEEEbiography}



\end{document}